%% file: gsm_main.tex
\documentclass[10pt,twocolumn,letterpaper]{article}

\usepackage{cvpr}
\usepackage{times}
\usepackage{epsfig}
\usepackage{graphicx}
\usepackage{amsmath}
\usepackage{amssymb}
\usepackage{subcaption}
\usepackage{multirow}
\usepackage[export]{adjustbox}
\usepackage[nolist]{acronym}
\usepackage{units}

\usepackage{xcolor}

\usepackage[pagebackref=true,breaklinks=true,letterpaper=true,colorlinks,bookmarks=false]{hyperref}

\cvprfinalcopy 

\def\cvprPaperID{7287} 

\ifcvprfinal\fi

\input{sub/acros}

\newcommand{\sota}{state-of-the-art }

\begin{document}
	
	\title{Gate-Shift Networks for Video Action Recognition}
	
	
	\author{Swathikiran Sudhakaran$^{1}$, Sergio Escalera$^{2,3}$, Oswald Lanz$^{1}$\\ 
	$^{1}$Fondazione Bruno Kessler, Trento, Italy\\
	$^{2}$Computer Vision Center, Barcelona, Spain\\
	$^{3}$Universitat de Barcelona, Barcelona, Spain\\
	{\tt\small \{sudhakaran,lanz\}@fbk.eu, \tt\small sergio@maia.ub.es}
}
	
	\maketitle
	\thispagestyle{empty}
	
	\begin{abstract}
Deep 3D CNNs for video action recognition are designed to learn powerful representations in the joint spatio-temporal feature space. In practice however, because of the large number of parameters and computations involved, they may under-perform in the lack of sufficiently large datasets for training them at scale. In this paper we introduce spatial gating in spatial-temporal decomposition of 3D kernels. We implement this concept with Gate-Shift Module (GSM). GSM is lightweight and turns a 2D-CNN into a highly efficient spatio-temporal feature extractor. With GSM plugged in, a 2D-CNN learns to adaptively route features through time and combine them, at almost no additional parameters and computational overhead. We perform an extensive evaluation of the proposed module to study its effectiveness in video action recognition, achieving state-of-the-art results on Something Something-V1 and Diving48 datasets, and obtaining competitive results on EPIC-Kitchens with far less model complexity.
\end{abstract}
	
\section{Introduction}
	\label{sec:intro}
	\input{sub/1_intro.tex}

\section{Related Work}
    \label{sec:related_works}
    \input{sub/2_related.tex}

\section{Gate-Shift Networks}
    \label{sec:gsm}
    \input{sub/3_gsm.tex}

\section{Experiments and Results}
    \label{sec:experiments}
    \input{sub/4_experiments.tex}

\section{Conclusion}
    \label{sec:conclusion}
	We proposed Gate-Shift Module (GSM), a novel temporal interaction block that turns a 2D-CNN into a highly efficient spatio-temporal feature extractor. GSM introduces spatial gating to decide on exchanging information with neighboring frames. We performed an extensive evaluation to study its effectiveness in video action recognition, achieving state-of-the-art results on Something Something-V1 and Diving48 datasets, and obtaining competitive results on EPIC-Kitchens with far less model complexity. For example, when GSM is plugged into TSN, an absolute gain of +32\% in recognition accuracy is obtained on Something Something-V1 dataset with just $0.48\%$ additional parameters and $0.55\%$ additional \acp{flop}.
	
	\section*{Acknowledgement}
	This work is partially supported by ICREA under the ICREA Academia programme.

	\newpage
{\small
\bibliographystyle{ieee_fullname}
\bibliography{gsm_bib}
}

\newpage
\section*{Appendix}
\appendix
\input{sub/5_supp.tex}

\end{document}

%% file: sub/acros.tex
\newacro{lstm}[LSTM]{Long Short-Term Memory}
\newacro{lsta}[LSTA]{Long Short-Term Attention}
\newacro{clstm}[ConvLSTM]{Convolutional Long Short-Term Memory}
\newacro{rnn}[RNN]{Recurrent Neural Network}
\newacro{cnn}[CNN]{Convolutional Neural Network}
\newacro{tsn}[TSN]{Temporal Segment Network}
\newacro{trn}[TRN]{Temporal Relation Network}
\newacro{tdn}[TDN]{Temporal Difference Network}
\newacro{mfn}[MFNet]{Motion Feature Network}
\newacro{fc}[FC]{Fully Connected}
\newacro{bn}[BN]{Batch Normalization}
\newacro{gru}[GRU]{Gated Recurrent Unit}
\newacro{sgd}[SGD]{Stochastic Gradient Descent}
\newacro{tsm}[TSM]{Temporal Shift Module}
\newacro{flop}[FLOP]{Floating Point Operation}
\newacro{lrcn}[LRCN]{Long-term Recurrent Convolutional Network}
\newacro{gsm}[GSM]{Gate-Shift Module}
\newacro{gsn}[GSN]{Gate-Shift Network}
\newacro{gst}[GST]{Grouped Spatial-temporal Aggregation}

%% file: sub/1_intro.tex
Video action recognition is receiving increasing attention due to its potential applications in video surveillance, media analysis, and robotics, just to mention a few. Although great advances have been achieved during last years, action recognition models have not yet achieved the success of image recognition models, and the `AlexNet momentum for video' has still to come. 

A key challenge lies in the space-time nature of the video medium that requires temporal reasoning for fine-grained recognition. 
Methods based on temporal pooling of frame-level features (TSN~\cite{tsn}, ActionVLAD~\cite{vlad}) process the video as a (order-less) set of still images and work well enough when the action can be discerned from objects and scene context (UCF-101, Sports-1M, THUMOS). More akin to the time dimension in video, late temporal aggregation of frame-level features can be formulated as sequence learning (LRCN~\cite{lrcn}, VideoLSTM~\cite{vlstm}) and with attention (Attentional Pooling~\cite{pool}, LSTA~\cite{lsta}). At the other hand, early temporal processing is used to fuse short term motion features from stack of flow fields (Two-Stream~\cite{twoStream}) or predicted directly from the encoded video (DMC-Net~\cite{dmc}). 

\begin{figure}
\includegraphics[width=.95\columnwidth]{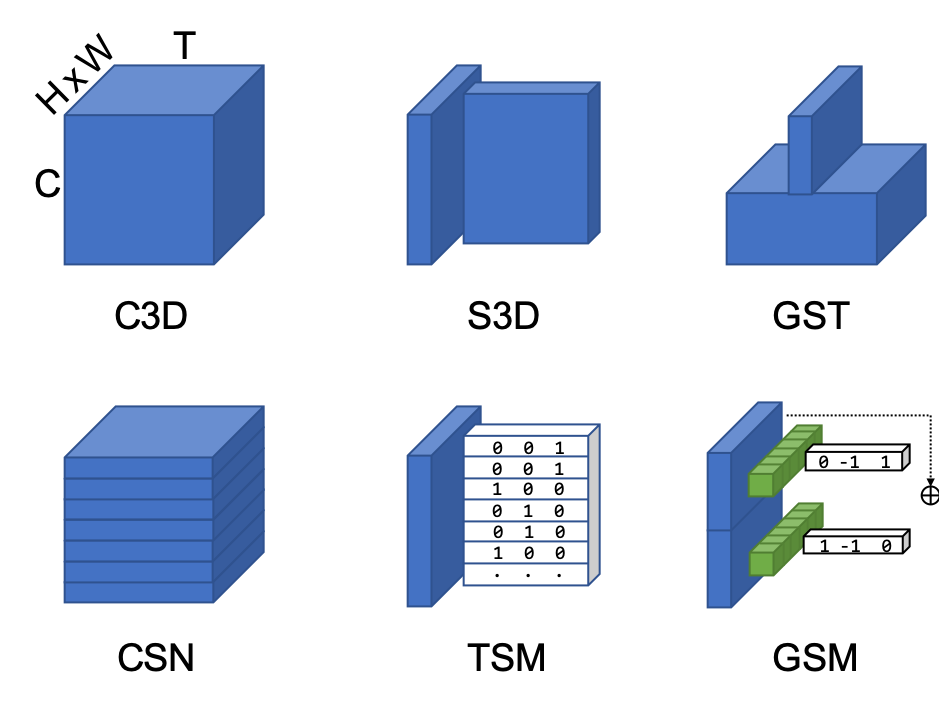}
\vspace{-0.3cm}
\caption{3D kernel factorization for spatio-temporal learning in video. Existing approaches decompose into channel-wise (CSN), spatial followed by temporal (S3D, TSM), or grouped spatial and spatio-temporal (GST). In all these, spatial, temporal, and channel-wise interaction is hard-wired. Our {\bf Gate-Shift Module (GSM)} leverages group spatial gating (blocks in green) to control interactions in spatial-temporal decomposition. GSM is lightweight and a building block of high performing video feature extractors.
}
\label{fig:kernel}
\end{figure}

Fine-grained recognition can benefit from deeper temporal modeling. Full-3D CNNs (C3D~\cite{tran2015learning,hara2018can}) process the video in space-time by
expanding the kernels of a 2D ConvNet along the temporal dimension. Deep C3Ds are designed to learn powerful representations in the joint spatio-temporal feature space with more parameters (3D kernels) and computations (kernels slide over 3 densely sampled dimensions). In practice however, they may under-perform due to the lack of sufficiently large datasets for training them at scale. To cope with these issues arising from the curse of dimension one can narrow down network capacity by design. Fig.~\ref{fig:kernel} shows several C3D kernel decomposition approaches proposed for spatio-temporal feature learning in video. A most intuitive approach is to factorize 3D spatio-temporal kernels into 2D spatial plus 1D temporal, 
resulting in a structural decomposition that disentangles spatial from temporal interactions (P3D~\cite{p3d}, R(2+1)D~\cite{r2plus1d}, S3D~\cite{s3d}). An alternative design is separating channel interactions and spatio-temporal interactions via group convolution (CSN~\cite{csn}), or modeling both spatial and spatio-temporal interactions in parallel with 2D and 3D convolution on separated channel groups (GST~\cite{gst}). Temporal convolution can be constrained to hard-coded time-shifts that move some of the channels forward in time or backward (TSM~\cite{tsm}). All these existing approaches learn structured kernels with a hard-wired connectivity and propagation pattern across the network. There is no data dependent decision taken at any point in the network to route features selectively through different branches, for example, group-and-shuffle patterns are fixed by design and learning how to shuffle is combinatorial complexity.

In this paper we introduce spatial gating in spatial-temporal decomposition of 3D kernels. We implement this concept with Gate-Shift Module (GSM) as shown in Fig.~\ref{fig:kernel}. GSM is lightweight and turns a 2D-CNN into a highly efficient spatio-temporal feature extractor. The GSM first applies 2D convolution, then decomposes the output tensor using a learnable spatial gating into two tensors: a gated version of it, and its residual. The gated tensor goes through a 1D temporal convolution while its residual is skip-connected to its output. We implement spatial gating as group spatio-temporal convolution with single output plane per group. We use hard-coded time-shift of channel groups instead of learnable temporal convolution. With GSM plugged in, a 2D-CNN learns to adaptively route features through time and combine them, at almost no additional parameters and computational overhead. For example, when GSM is plugged into TSN~\cite{tsn}, an absolute gain of $+32$ percentage points in accuracy is obtained on Something Something-V1 dataset with just $0.48\%$ additional parameters and $0.55\%$ additional \acp{flop}. 

The contributions of this paper can be summarized as follows: (i) We propose a novel spatio-temporal feature extraction module that can be plugged into existing 2D \ac{cnn} architectures with negligible overhead in terms of computations and memory; (ii) We perform an extensive ablation analysis of the proposed module to study its effectiveness in video action recognition; (iii) We achieve \sota or competititve results on public benchmarks with less parameters and \acp{flop} compared to existing approaches.

%% file: sub/2_related.tex
Inspired by the performance improvements obtained with deep convolutional architectures in image recognition~\cite{resnet, szegedy2016rethinking}, much effort has gone into extending these for video action recognition.

\vspace*{4pt}
\noindent{\bf Fusing appearance and flow.} 
A popular extension of 2D CNNs to handle video is the Two-Stream architecture by Simonyan and Zisserman~\cite{twoStream}. Their method consists of two separated CNNs (streams) that are trained to extract features from a sampled RGB video frame paired with the surrounding stack of optical flow images, followed by a late fusion of the prediction scores of both streams. The image stream encodes the appearance information while the optical flow stream encodes the motion information, that are often found to complement each other for action recognition. 
Several works followed this approach to find a suitable fusion of the streams at various depths~\cite{feichtenhofer2016convolutional} and to explore the use of residual connections between them~\cite{feichtenhofer2016spatiotemporal}. These approaches rely on optical flow images for motion information, and a single RGB frame for appearance information, which is limiting when reasoning about the temporal context is required for video understanding.

\vspace*{4pt}
\noindent{\bf Video as a set or sequence of frames.} Later, other approaches were developed using multiple RGB frames for video classification. These approaches sparsely sample multiple frames from the video, which are applied to a 2D \ac{cnn} followed by a late integration of frame-level features using average pooling~\cite{tsn}, multilayer perceptrons~\cite{trn}, recurrent aggregation~\cite{lrcn, vlstm}, or attention~\cite{pool,lsta}. To boost performance, most of these approaches also combine video frame sequence with externally computed optical flow. This shows to be helpful, but computationally intensive.

\vspace*{4pt}
\noindent{\bf Modeling short-term temporal dependencies.}
Other research has investigated the middle ground between late aggregation (of frame features) and early temporal processing (to get optical flow), by modeling short-term dependencies.
This includes differencing of intermediate features~\cite{tdn} and combining Sobel filtering with feature differencing~\cite{off}. Other works~\cite{tvnet, piergiovanni2019representation} develop a differentiable network that performs TV-L1~\cite{tv-l1}, a popular optical flow extraction technique. The work of \cite{mfnet} instead uses a set of fixed filters for extracting motion features, thereby greatly reducing the number of parameters. DMC-Nets~\cite{dmc} leverage motion vectors in the compressed video to synthesize discriminative motion cues for two-stream action recognition at low computational cost compared to raw flow extraction.

\vspace*{4pt}
\noindent{\bf Video as a space-time volume.} Unconstrained modeling and learning of action features is possible when considering video in space-time.
Since video can be seen as a temporally dense sampled sequence of images, expanding 2D convolution operation in 2D-CNNs to 3D convolution is a most intuitive approach to spatio-temporal feature learning~\cite{tran2015learning, hara2018can, carreira2017quo}. The major drawback of 3D \acp{cnn} is the huge number of parameters involved. This results in increased computations and the requirement of large scale datasets for pre-training. Carreira and Zisserman~\cite{carreira2017quo} addressed this limitation by inflating video 3D kernels with the 2D weights of a \ac{cnn} trained for image recognition. Several other approaches focused on reducing the number of parameters by disentangling the spatial and temporal feature extraction operations.
P3D~\cite{p3d} proposes three different choices for separating the spatial and temporal convolutions and develops a 3D-ResNet architecture whose residual units are a sequence of such three modules. R(2+1)D~\cite{r2plus1d} and S3D-G~\cite{s3d} also show that a 2D convolution followed by 1D convolution is enough to learn discriminative features for action recognition.
CoST~\cite{cost} performs 2D convolutions, with shared parameters, along the three orthogonal dimensions of a video sequence. MultiFiber~\cite{multifiber} uses multiple lightweight networks, the fibers, and multiplexer modules that facilitate information flow using point-wise convolutions across the fibers.

\vspace*{4pt}
\noindent{\bf Spatial-temporal modeling.} 
Recently, the focus of research is moving to the development of efficient (from a computational point of view) and effective (from a performance point of view) architectures. 
\acp{cnn} provide different levels of feature abstractions at different layers of the hierarchy. It has been found that the bottom layer features are less useful for extracting discriminative motion cues~\cite{sun2015human, eco, s3d}. In \cite{sun2015human} it is proposed to apply 1D convolution layers on top of a 2D \ac{cnn} for video action recognition. The works of \cite{eco} and \cite{s3d} show that it is more effective to apply full 3D and separable 3D convolutions at the top layers of a 2D \ac{cnn} for extracting spatio-temporal features. 
These approaches resulted in performance improvement over full 3D architectures with less parameters and computations. Static features from individual frames represent scenes and objects and can also provide important cues in identifying the action. This is validated by the improved performance obtained with two-path structures that apply a parallel 2D convolution in addition to the 3D convolution~\cite{mict, gst}. MiCT~\cite{mict} is designed by adding 3D convolution branches in parallel to the 2D convolution branches of a BN-Inception-like \ac{cnn}. GST~\cite{gst} makes use of the idea of grouped convolutions for developing an efficient architecture for action recognition. They separate the features at a hierarchy across the channel dimension and separately perform 2D and 3D convolutions followed by a concatenation operation. In this way, the performance is increased while reducing the number of parameters. STM~\cite{stm} proposes two parallel blocks for extracting motion features and spatio-temporal features. Their network rely only on 2D and 1D convolutions and feature differencing for encoding motion and spatio-temporal features. TSM~\cite{tsm} proposes to shift the features across the channel dimension as a way to perform temporal interaction between the features from adjacent frames of a video. This parameter-less approach has resulted in similar performance to 3D \acp{cnn}. However, in all previous approaches, spatial, temporal, and channel-wise  interaction is hard-wired. Here, we propose the Gate-Shift  Module  (GSM), which control interactions in spatial-temporal decomposition and learns to adaptively route features thought time and combine them, at almost no additional parameters and computational overhead.

%% file: sub/3_gsm.tex
In this section we present Gate-Shift Networks for fine-grained action recognition. We first describe their building block, \ac{gsm}, that turns a 2D~\ac{cnn} into a high performing spatio-temporal feature extractor, with minimal overhead. We then discuss and motivate the design choices leading to our final GSM architecture used in the experiments.

\begin{figure*}
    \centering\includegraphics[width=0.8\linewidth]{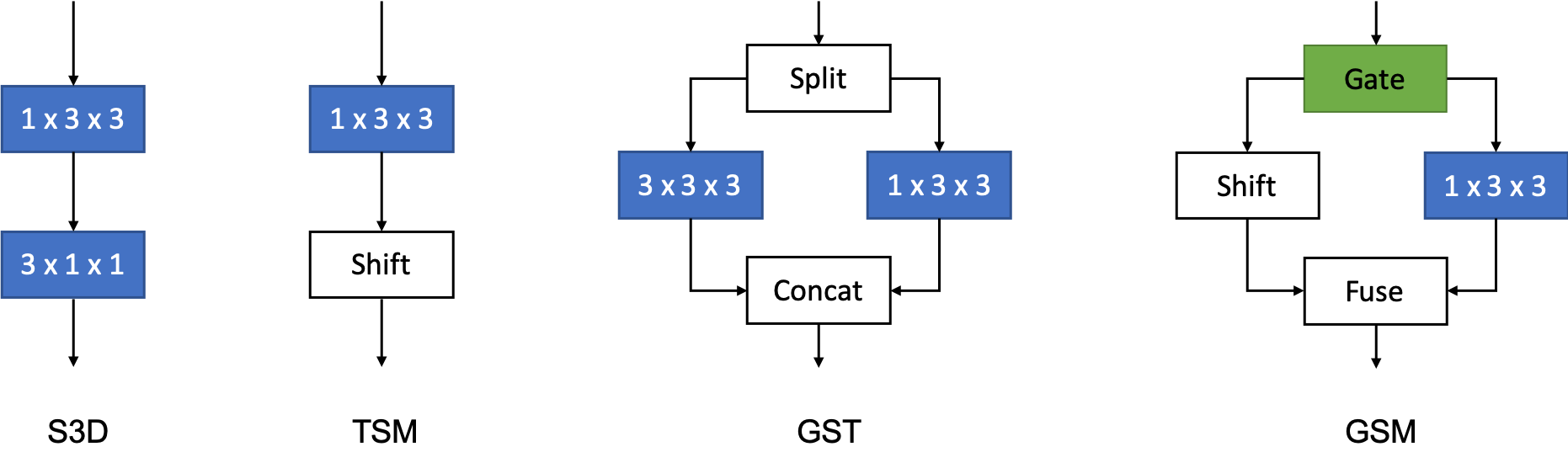}\vspace{-0.4cm}
	\caption{C3D  decomposition approaches in comparison to GSM schematics. GSM is inspired by GST and TSM but replaces the hard-wired channel split with a learnable spatial gating block. 
	}
	\label{fig:comp}
\end{figure*}

\subsection{\acl{gsm}}
\label{sec:gsm_module}

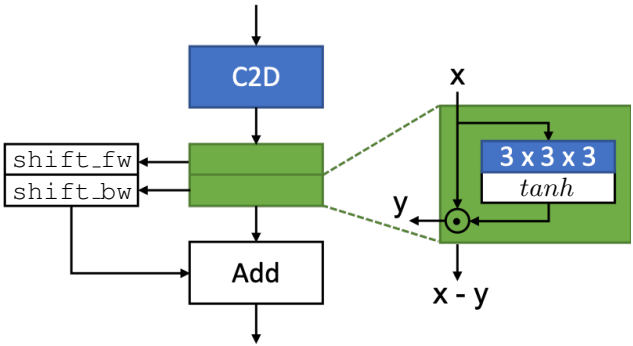
\begin{figure}
	\centering
\includegraphics[width=\columnwidth]{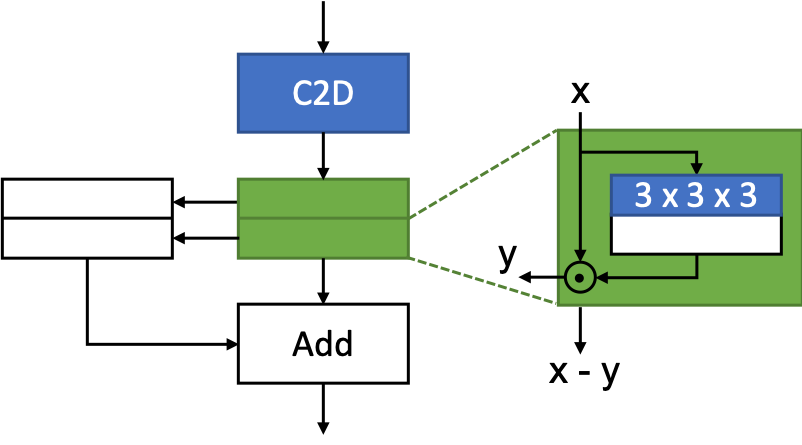}\begin{picture}(0,0)
\put(-234.2,69.7){\texttt{shift\_fw}}
\put(-234.2,57.8){\texttt{shift\_bw}}
\put(-42.,58.8){$tanh$}
\vspace{-0.4cm}
\end{picture}
	\caption{\ac{gsm} implementation with group gating and forward-backward temporal shift. A gate is a single 3D convolution kernel with tanh calibration, thus very few parameters are added when GSM is used to turn a C2D base model into a spatio-temporal feature extractor. 
	}
	\label{fig:gsm}
\end{figure}

Fig.~\ref{fig:comp} illustrates the network schematics of 3D kernel factorization approaches (cf. Fig.~\ref{fig:kernel}) that have been successfully applied to video action recognition. S3D, or R(2+1)D, P3D, decompose 3D convolutions into 2D spatial plus 1D temporal convolutions. \acs{tsm} replaces 1D temporal convolution with parameter-free channel-wise temporal shift operations. GST uses group convolution where one group applies 2D spatial and the other 3D spatio-temporal convolution. GST furthermore applies point-wise convolution before and after the block to allow for interactions between spatial and spatio-temporal groups, and for channel reduction and up-sampling. In these modules, the feature flow is hard-wired by design, meaning that features are forwarded from one block to the next without data-dependent pooling, gating or routing decision. 

GSM design, in Fig.~\ref{fig:comp}, is inspired by GST and TSM but replaces the hard-wired channel split with a learnable spatial gating block. The function of gate block, paired with fuse block, is selectively routing gated features through time-shifts and merging them with the spatially convolved residual to inject spatio-temporal interactions adaptively. GSM is lightweight as it uses 2D kernels, parameter-free time-shifts and few additional parameters to compute the spatial gating planes. 

Based on the conceptual design in Fig.~\ref{fig:comp}, we instantiate GSM as in Fig.~\ref{fig:gsm}. GSM first applies spatial convolution on the layer input; this is the operation inherited from the 2D CNN base model where GSM is build in. Then, grouped spatial gating is applied, that is, gating planes are obtained for each of two channel groups, and applied on them. This separates the 2D convolution output into group-gated features and residual. The gated features are group-shifted forward and backward in time, and zero-padded. These are finally fused (added) with the residual and propagated to the next layer. This way, GSM selectively mixes spatial and temporal information through a learnable spatial gating. 

Gating is implemented with a single spatio-temporal 3D kernel and tanh activation. With a 3D kernel we utilize short-range spatio-temporal information in the gating. tanh provides spatial gating planes with values in the range $(-1,+1)$ and is motivated as follows. When the gating value at a feature location is +1 and that of the time-shifted feature was +1, then a temporal feature averaging is performed at that location. If the gating value of the time-shifted feature was -1 instead, then a temporal feature differencing is performed. Using tanh, the gating can thus learn to apply either of the two modes, location-wise. It is also found in our ablation study that tanh provides better results than \eg sigmoid that would be the standard choice with gating, see Sec.~\ref{sec:ablation} last paragraph.

\vspace*{4pt}
\noindent\textbf{GSM layer implementation.}
Let tensor $X$ be the GSM input after 2D convolution (output of blue block in Fig.~\ref{fig:gsm}), of shape $C\times T\times W\times H$ where $C$ is the number of channels and $WH,T$ are the spatial and temporal dimensions. Let $X=[X_1,X_2]$ be the group=2 split of $X$ along the channel dimension, and $W=[W_1, W_2]$ be the two $\nicefrac{C}{2}\times 3\times 3\times 3$ shaped gating kernels. Then, the GSM output $Z = [Z_1, Z_2]$ is computed as
\begin{eqnarray}
 Y_1 &=& tanh(W_1 * X_1) \odot X_1 \label{eq:y1}\\
 Y_2 &=& tanh(W_2 * X_2) \odot X_2 \label{eq:y2}\\
 R_1 &=& X_1-Y_1 \label{eq:r1}\\
 R_2 &=& X_2-Y_2 \label{eq:r2}\\
 Z_1 &=& \verb+shift_fw+(Y_1) + R_1 \label{eq:z1}\\
 Z_2 &=& \verb+shift_bw+(Y_2) + R_2 \label{eq:z2}
 \end{eqnarray}
where `$*$' represents convolution, `$\odot$' is Hadamard product, and $\verb+shift_fw+,\verb+shift_bw+$ is forward, backward temporal shift. Note that the parameter count in this GSM implementation is $2\times (27\cdot \nicefrac{C}{2}) = 27\cdot C$; this is far less than that of a typical C2D block. E.g., the $1\times 3\times 3$ block in Fig.~\ref{fig:gsm} has $C$ kernels of size $(9\cdot C_{in})$ where typically $C\ge C_{in} \gg 3$.

\vspace*{4pt} 
\noindent\textbf{Relation to Residual Architectures.}
It should be noted that Eqns.~\ref{eq:r1} and~\ref{eq:z1} can be reformulated as $R_1=\texttt{shift\_fw}(Y_1)-Y_1$ and $Z_1=X_1 + R_1$. This is in analogy to the residual learning of ResNet. In GSM, the residual is the learned spatio-temporal features that are added to the input $X_1$ to generate discriminative spatio-temporal features for identifying an action class.

\vspace*{4pt} 
\noindent\textbf{Relation to existing approaches.}
GSM is a generalization of several existing approaches. With gating=0, GSM collapses to TSN~\cite{tsn}; with gating=1, converges to TSM~\cite{tsm} style; with gating=1 and replacing temporal shift with expensive 3D convolution, converges to GST~\cite{gst} style.

\subsection{Gate-Shift Architecture}
\label{sec:gsn}
We adopt TSN as reference architecture for action recognition. TSN performs temporal pooling of frame-level features using C2D backbone. We choose BN-Inception and InceptionV3 as backbone options with TSN, and describe here how we GSM them.

\begin{figure}
	\begin{subfigure}[b]{0.45\columnwidth}
		\centering
		\includegraphics[scale=0.1]{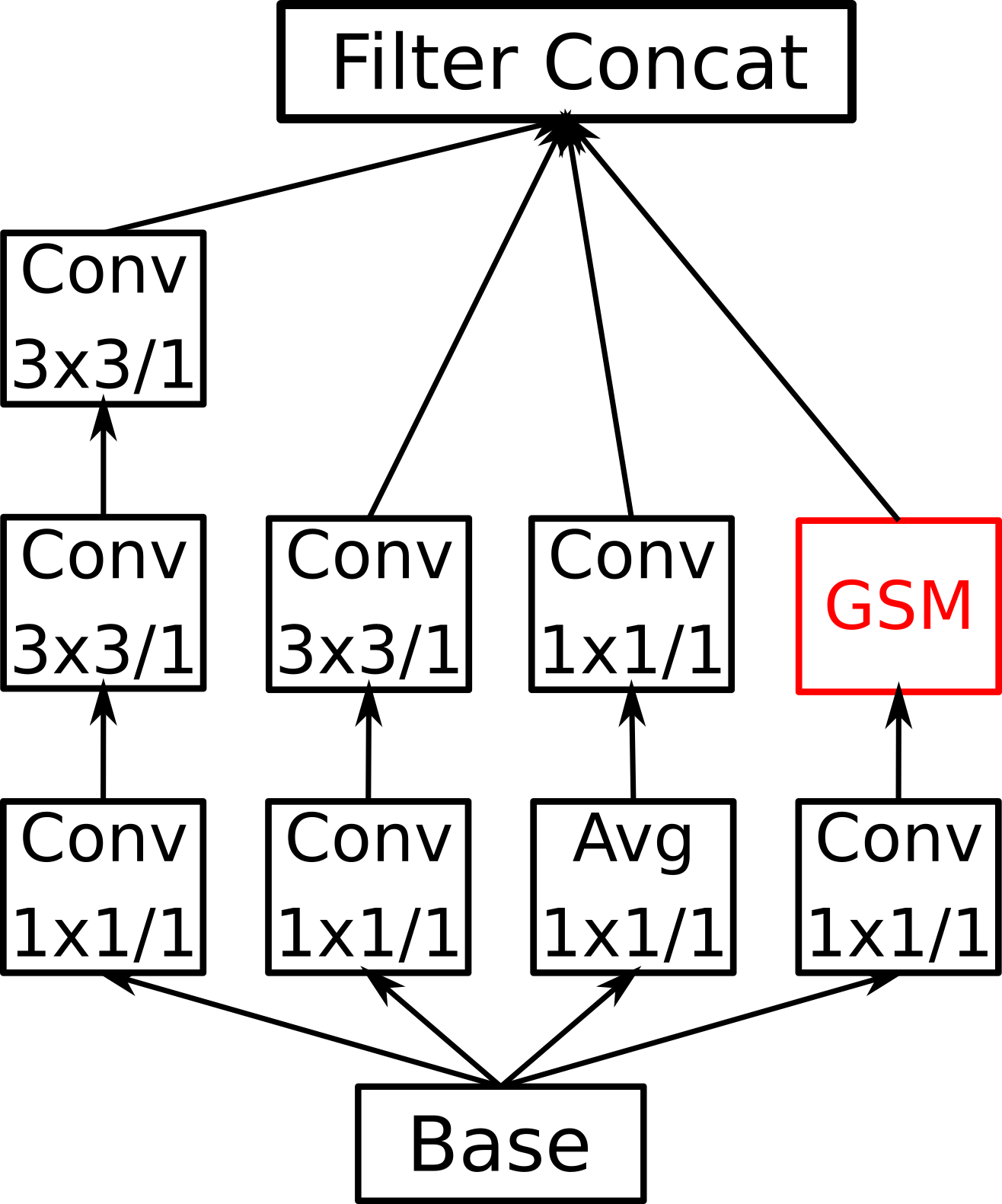}
	\end{subfigure} \hfill
	\begin{subfigure}[b]{0.45\columnwidth}
		\centering
		\includegraphics[scale=0.1]{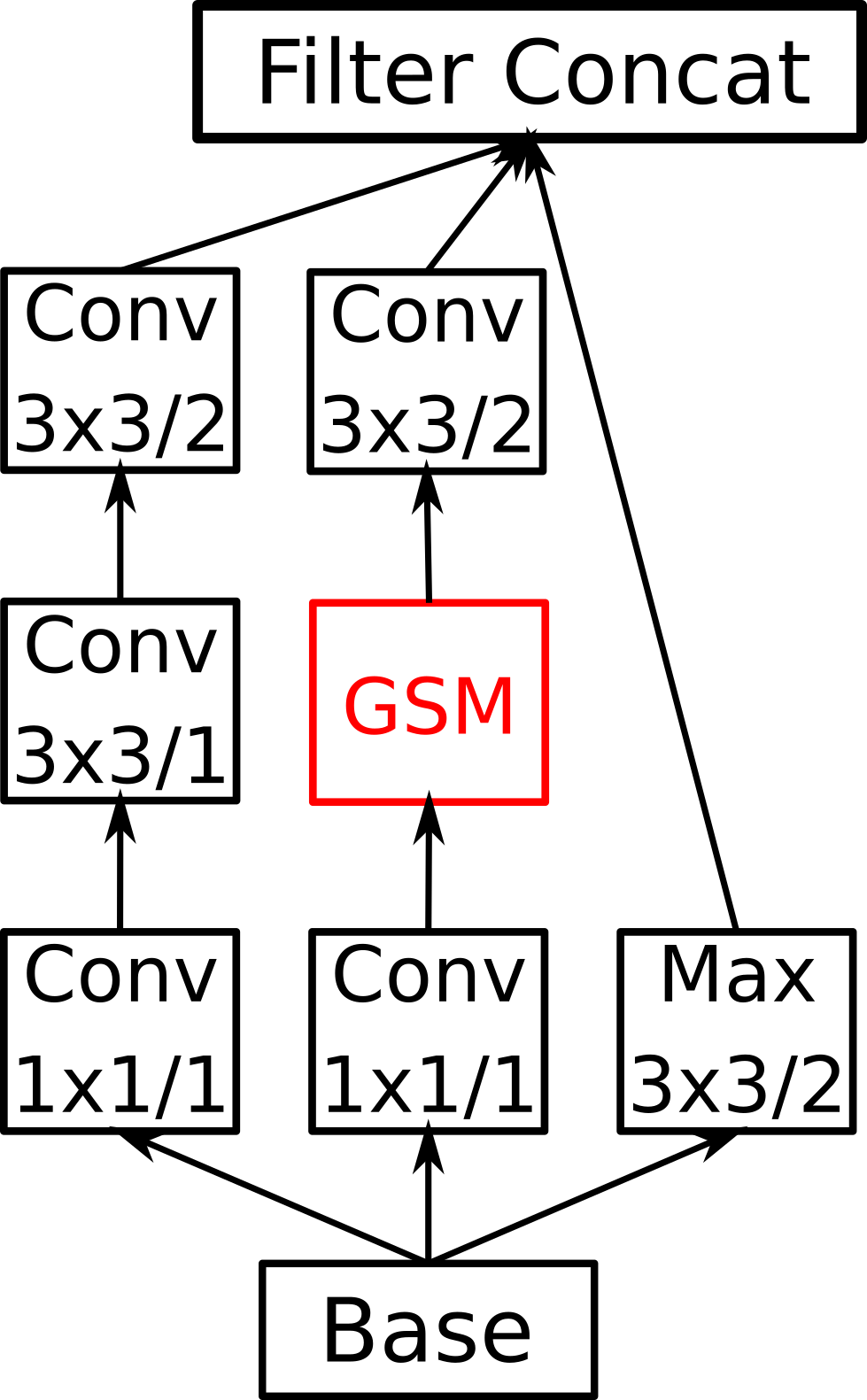}
	\end{subfigure}
	\caption{BN-Inception blocks with GSM. The kernel size and stride of convolutional and pooling layers are annotated inside each block.
	}
	\label{fig:gsn_blocks}
\end{figure}

As shown in Fig.~\ref{fig:gsn_blocks}, we insert GSM inside one of the branches of Inception blocks. We analyze the branch to which \ac{gsm} is to be applied, empirically. From the experiments we conclude that adding GSM to the branch with the least number of convolution layers performs the best. A hypothesis is that 
the other branches consist of spatial convolutions with larger kernel sizes and applying \ac{gsm} on those branches will affect the spatial learning capacity of the network. This hypothesis is strengthened further by observing a reduced performance when \ac{gsm} was added inside all the branches. Because of the presence of additional branches in Inception blocks, that encode spatial information, there is no need for a separate spatial convolution operation in \ac{gsm}. That is, the GSM blocks in Fig.~\ref{fig:gsn_blocks} are as in Fig.~\ref{fig:gsm} without the $1\times 3\times 3$ spatial convolution block.

For clip level action classification we follow the approach of TSN, that is, we predict the action by average pooling the frame level (now spatio-temporal) scores.

%% file: sub/4_experiments.tex
This section presents an extensive set of experiments to evaluate \ac{gsm}.

\begin{figure*}[h!]
	\begin{subfigure}[b]{0.31\textwidth}
	\centering
		\includegraphics[scale=1]{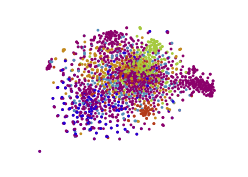}
	    \vspace{-0.7cm}
		\caption{t-SNE plot of TSN features}
		\label{fig:tsn_tsne}
	\end{subfigure}\hfill
	\begin{subfigure}[b]{0.31\textwidth}
	\centering
	    \includegraphics[scale=1]{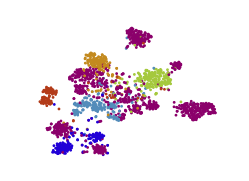}
	    \vspace{-0.7cm}
		\caption{t-SNE plot of GSM features}
		\label{fig:hftsn_tsne}
	\end{subfigure}\hfill
	\begin{subfigure}[b]{0.31\textwidth}
	\centering
	    \includegraphics[scale=1]{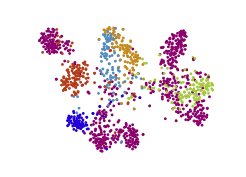}
	    \vspace{-0.7cm}
		\caption{t-SNE of most improved classes}
		\label{fig:hftsn_tsne_improved}
	\end{subfigure}\\[.2in]
	\centering
	\begin{subfigure}[b]{\textwidth}
		\centering
		\includegraphics[angle=90, scale=0.3]{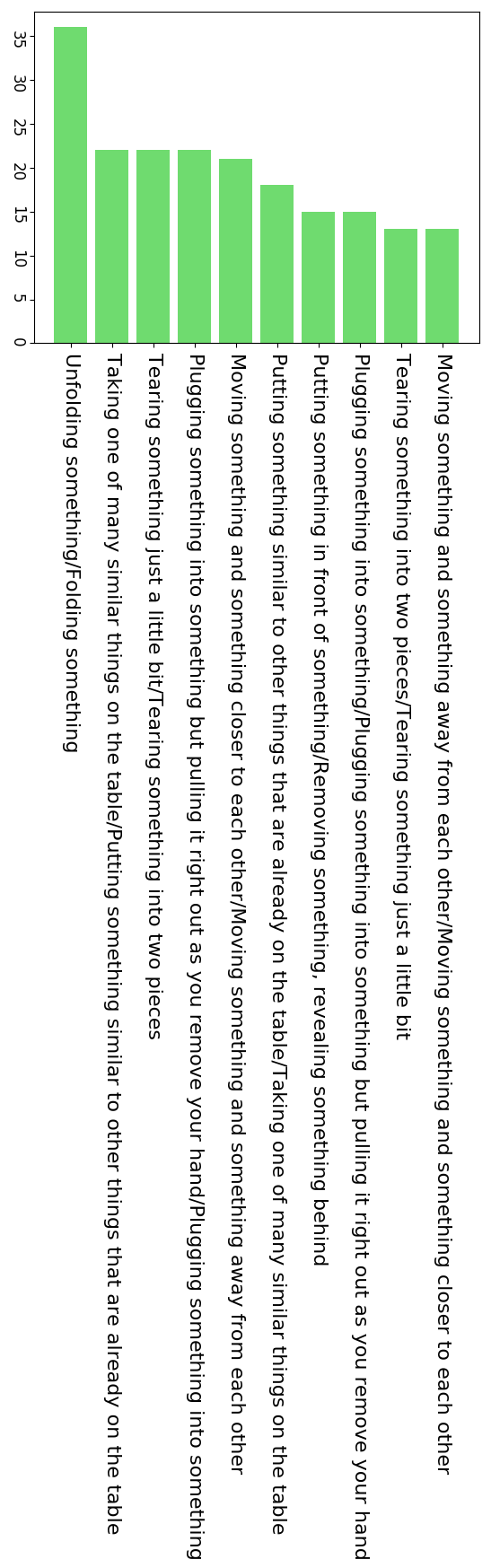}
	    \vspace{-0.4cm}
		\caption{Most improved classes}
		\label{fig:classes_imp}
	\end{subfigure}
    \caption{t-SNE plots of the output layer features preceding the final fully connected layers for (a) TSN with BN-Inception, and (b) same TSN but with GSM built-in as in Fig.~\ref{fig:gsn_blocks}. In these two plots the 10 action categories described in~\cite{goyal2017something} are visualized. In (d) we list the action classes with the highest improvement over \ac{tsn} baseline. X-axis shows the number of corrected samples for each class. Y-axis labels are in the format true label (\ac{gsm})/predicted label (\ac{tsn}). In (c) we visualize the corresponding t-SNE plot.
    }
	\label{fig:tsne_plots}
\end{figure*}
\subsection{Datasets}

We evaluate \acf{gsm} on three standard action recognition benchmarks, Something Something-V1~\cite{goyal2017something} (Something-V1), Diving48~\cite{diving} and EPIC-Kitchens~\cite{damen2018scaling}.  Something-V1 consists of 100K videos with 174 fine-grained object manipulation actions. Performance is reported on the validation set. Diving48 dataset contains around 18K videos with 48 fine-grained dive classes. EPIC-Kitchens dataset comprises 34K egocentric videos with fine-grained activity labels. We report the performance obtained on the two standard test splits. Since the test labels are withheld, the recognition scores are obtained from the submission server after we submitted the prediction scores.

All the three considered datasets are diverse in nature and require strong spatio-temporal reasoning for predicting the action classes. For instance, Something-V1 dataset does not distinguish among the objects being handled. On the other hand, EPIC-Kitchens dataset require strong spatio-temporal reasoning as well as information about the objects being handled. The videos in Diving48 dataset generally contain a uniform background with fine-grained diving actions and require strong understanding of temporal dynamics of the human body in the video. 

\subsection{Implementation Details}

As explained in Sec.~\ref{sec:gsn}, we choose BN-Inception and InceptionV3 as the \ac{cnn} backbones. \ac{gsm} is added inside each Inception block of the respective \acp{cnn}. Thus a total of 10 \acp{gsm} are added. We initialize the 3D convolution in the gating layer with zeros. Thus the model starts as a standard \ac{tsn} architecture and the gating is learned during training. All models are initialized with ImageNet pre-trained weights. The entire network is trained end-to-end using \ac{sgd} with an initial learning rate of 0.01 and momentum 0.9. We use a cosine learning rate schedule~\cite{sgdr}. The network is trained for 60 epochs on Something-V1 and EPIC-Kitchens while Diving48 is trained for 20 epochs. The first 10 epochs are used for gradual warm-up~\cite{goyal2017accurate}. The batch size is 32 for Something-V1 and EPIC-Kitchens, and 8 for Diving48. Dropout is applied at the classification layer at a rate of 0.5 for Something-V1 and EPIC-Kitchens and 0.7 for Diving48 dataset. Random scaling, cropping and flipping are applied as data augmentation during training. The dimension of the input is $224\times224$ and $229\times229$ for BN-Inception and InceptionV3, respectively. The reduced input dimension to InceptionV3 reduces the computational complexity without degradation in performance. If not specified, we use just the center crop during inference.

\begin{table}[t]
\centering\renewcommand{\arraystretch}{0.7}
	\begin{tabular}{|c|c|}
		\hline
		\textbf{Branch} & \textbf{Accuracy (\%)} \\ \hline \hline
		Branch 1 & 45.11 \\ \hline
		Branch 2 & 44.98 \\ \hline
		Branch 3 & 45.05 \\ \hline
		\textbf{Branch 4} & \textbf{47.24} \\ \hline
		All branches & 43.5 \\ \hline
	\end{tabular}
	\caption{Ablation analysis done to determine the Inception branch that is most suitable for plugging in \ac{gsm}.}
	\label{tab:ablation_branch}
\end{table}

\begin{table}[t]\small\centering\setlength\tabcolsep{4.5pt}\renewcommand{\arraystretch}{0.7}
	\begin{tabular}{|c|c|c|c|}
		\hline
		\textbf{Model} & \textbf{Accuracy (\%)} & \textbf{Params.} & \textbf{FLOPs}\\ \hline \hline
		BN-Inception (baseline) & 17.25 & 10.45M & 16.37G\\ \hline
		BN-Inception + 1 GSM & 22.7 & 10.46M & 16.37G\\ \hline
		BN-Inception + 5 GSM & 43.13 & 10.48M & 16.39G\\ \hline
		\textbf{BN-Inception + 10 GSM} & \textbf{47.24} & \textbf{10.5M} & \textbf{16.46G}\\ \hline
	\end{tabular}
	\caption{Recognition Accuracy by varying the number of \acfp{gsm} added to the backbone.}
	\label{tab:ablation}
\end{table}

\subsection{Ablation Analysis}
\label{sec:ablation}
In this section, we report the ablation analysis performed on the validation set of Something-V1 dataset. In all the experiments, we apply 8 frames as input to the network. We first conducted an analysis to determine the Inception branch that is most suitable for adding \ac{gsm}. The results of this experiment are reported in Tab.~\ref{tab:ablation_branch}. We number each branch from left to right. It can be seen that the best performing model is obtained when \ac{gsm} is added in branch 4. When \ac{gsm} is added inside all the branches, we observed the lowest performance as this adversely affects the spatial modeling capacity of the network. From the above experiments, we conclude that \ac{gsm} is most suited to be added inside the branch which contains the least number of convolutions. We follow the same design choice for InceptionV3 as well. More details regarding the architecture of InceptionV3 is provided in the supplementary material.

We then compared the performance improvement by adding \ac{gsm} on BN-Inception. Tab.~\ref{tab:ablation} shows the ablation results. Baseline is the standard \ac{tsn} architecture, with an accuracy of $17.25\%$. We then applied \ac{gsm} at the last Inception block of the \ac{cnn}. This improved the recognition performance by $5\%$. Increasing the number of \acp{gsm} added to the backbone consistently improved the recognition performance of the network. The final model, in which \ac{gsm} is applied in all Inception blocks results in a recognition accuracy of $47.24\%$, \ie, $+30\%$ absolute improvement over \ac{tsn} baseline, with only $0.48\%$ and $0.55\%$ overhead in parameters and complexity, respectively.

Since sigmoid is the general choice used in gating mechanims, we also analyzed the performance of \ac{gsm} when sigmoid non-linearity is used inside the gating function. Compared to tanh non-linearity, sigmoid unperforms by absolute $3\%$ (47.24\% vs 44.75\%) proving the suitability of tanh for gate calibration.

\subsection{State-of-the-Art Comparison}

\begin{table*}[t]\small
	\centering\renewcommand{\arraystretch}{0.5}
	\begin{tabular}{|c|c|c|c|c|c|}
		\hline
		\textbf{Method} & \textbf{Backbone} & \textbf{Pre-training} & \textbf{\#Frames} & \textbf{GFLOPs} & \textbf{Accuracy (\%)} \\ \hline \hline
		\ac{tsn}~\cite{tsn} (ECCV'16) & BN-Inception & ImageNet & 16 & 32.73 & 17.52 \\ \hline
		MultiScale \acs{trn}~\cite{trn} (ECCV'18) & BN-Inception & ImageNet & 8 & 16.37 & 34.44  \\ \hline
		R(2+1)D~\cite{r2plus1d} (CVPR'18) & ResNet-34 & Sports-1M & 32 & 152 & 45.7  \\ \hline
		R(2+1)D~\cite{r2plus1d} from~\cite{ghadiyaram2019large} (CVPR'19) & ResNet-34 & External & 32 & 152 & 51.6 \\ \hline
		S3D-G~\cite{s3d} (ECCV'18) & InceptionV1 & ImageNet & 64 & 71.38 & 48.2 \\ \hline
		MFNet~\cite{mfnet} (ECCV'18) & ResNet-101 & - & 10 & NA & 43.9 \\ \hline
		TrajectoryNet~\cite{trajectoryNet} (NeurIPS'18) & ResNet-18 & Kinetics & 7$\times$10 & NA & 47.8 \\ \hline
		\acs{tsm}~\cite{tsm} (ICCV'19) & ResNet-50 & Kinetics & 16 & 65 & 47.2  \\ \hline
		STM~\cite{stm} (ICCV'19) & ResNet-50 & ImageNet & 16$\times$30 & 66.5$\times$30 & 50.7  \\ \hline
		\acs{gst}~\cite{gst} (ICCV'19) & ResNet-50 & ImageNet & 16 & 59 & 48.6 \\ \hline
		ABM~\cite{abm} (ICCV'19) & ResNet-50 & ImageNet & 16$\times$3 & 35.33$\times$3 & 46.08 \\ \hline
		CorrNet~\cite{corrnet} & ResNet-101 & - & 32$\times$30 & 224$\times$30 & 51.1 \\ \hline \hline	
		I3D~\cite{carreira2017quo} (CVPR'17) & ResNet-50 & Kinetics & 32$\times$2 & 108$\times$2 & 41.6 \\ \hline
		Non-local~\cite{nonlocal} (CVPR'18) & ResNet-50 & Kinetics & 32$\times$2 & 168$\times$2 & 44.4 \\ \hline
		GCN+Non-local~\cite{wang2018videos} (ECCV'18) & ResNet-50 & Kinetics & 32$\times$2 & 303$\times$2 & 46.1 \\ \hline
		ECO~\cite{eco} (ECCV'18) & BNInc + ResNet-18 & Kinetics & 16 & 64 & 41.4  \\ \hline
		Martinez \etal~\cite{martinez2019action} (ICCV'19) & ResNet-50 & ImageNet & NA & 52.17$^*$$\times$NA & 50.1 \\ \hline
		Martinez \etal~\cite{martinez2019action} (ICCV'19) & ResNet-152 & ImageNet & NA & 113.4$^*$$\times$NA & 53.4 \\ \hline \hline
		\multirow{5}{*}{\ac{gsm}} 
		 & BN-Inception & ImageNet & 8 & 16.46 & 47.24  \\ \cline{2-6}
		 & InceptionV3  & ImageNet & 8 & 26.85 & 49.01  \\ \cline{2-6}
    	 & BN-Inception & ImageNet & 16 & 32.92 & 49.56 \\ \cline{2-6}
		 & InceptionV3  & ImageNet & 16 & 53.7 & 50.63 \\ \cline{2-6}
		 & InceptionV3  & ImageNet & 16$\times$2 & 53.7$\times$2 & 51.68 \\ \cline{2-6}
		 & InceptionV3 & ImageNet & 8$\times$2+12$\times$2+16+24 & 268.47 & 55.16\\ \hline
	\end{tabular}
	\caption{Comparison to \sota on Something-V1. $^*$: Computed assuming a single clip of 16 frames as input.}
	\label{tab:something_sota}
\end{table*}

\textbf{Something-V1.} The recognition performance obtained by \ac{gsm} is compared with \sota approaches that just use RGB frames in Tab.~\ref{tab:something_sota}. We also report the number of frames used by each approach during inference and the corresponding computational complexity in terms of \acp{flop}. The first block in the table lists the approaches that use 2D~\ac{cnn} and efficient 3D~\ac{cnn} implementations. The second block shows the approaches that use Full-3D~\acp{cnn}. From the table, it can be seen that \ac{gsm} results in an absolute gain of $+32\%$ ($17.52\%$ vs $49.56\%$) over the \ac{tsn} baseline. \ac{gsm} performs better than 3D~\acp{cnn} or heavier backbones and also those approaches that use external data for pre-training, with considerably less number of \acp{flop}. \ac{gsm} performs comparatively to the top performing method~\cite{martinez2019action} with less number of \acp{flop}. It should be noted that the \acp{flop} of the architecture described in~\cite{martinez2019action} is computed assuming a single clip of 16 frames. It can also be seen that by using InceptionV3, which is a larger backbone than BN-Inception, the performance of the proposed approach improves. We also evaluated the performance of ensemble of models trained with different number of input frames and achieved a state-of-the-art recognition accuracy of 55.16\% \footnote{See supplementary document for more details on model ensembles.}.

\textbf{Diving48.}
Tab.~\ref{tab:diving_sota} compares performance of \ac{gsm} on Diving48 dataset  with \sota approaches. We train the network using 16 frames and sample two clips during inference. We use InceptionV3 as the \ac{cnn} backbone. In this dataset, the actions cannot be recognized from the scene context alone and require strong spatio-temporal reasoning. \ac{gsm} achieves a recognition accuracy of $40.27\%$, an improvement of $+1.3\%$ over previous \sota~\cite{gst}.

\textbf{EPIC-Kitchens.}
In EPIC-Kitchens, the labels are provided as \verb|verb|-\verb|noun| pairs and the performance is evaluated on \verb|verb|, \verb|noun| and \verb|action| recognition accuracies. For this dataset, we train \ac{gsm} as a multi-task problem for \verb|verb|, \verb|noun| and \verb|action| prediction. In the classification layers, we apply \verb|action| scores as bias to the \verb|verb| and \verb|noun| classifiers, as done in \acs{lsta}~\cite{lsta}. We use BN-Inception as the backbone \ac{cnn}. The network is trained with 16 frames. Two clips consisting of 16 frames are sampled from each video during inference. We report the recognition accuracy obtained on the two standard test splits, S1 (seen) and S2 (unseen), in Tab.~\ref{tab:epic_sota}. The first block in the table shows the methods that use both RGB frames and optical flow as inputs while the second block lists the approaches that only use RGB images. From the table, it can be seen that \ac{gsm} performs better than other approaches that use optical flow images for explicit motion encoding. The only two methods that beat \ac{gsm}, R(2+1)D~\cite{ghadiyaram2019large} and LFB~\cite{lfb}, train two separate networks, one trained for \verb|verb| and the other for \verb|noun| classification, and leverage additional data for pre-training. \ac{gsm} uses a single network for predicting all three labels from a video, thereby making it faster and more memory efficient. In fact, \ac{gsm} performs better than R(2+1)D model pre-trained on Sports-1M dataset, suggesting that \ac{gsm} can also improve its performance by pre-training on external data.

\subsection{Discussion}

\begin{figure}
	\captionsetup{font=scriptsize, labelfont=scriptsize}
	\centering
	\begin{subfigure}[b]{0.49\linewidth}\centering
		\captionsetup{labelformat=empty}
		\includegraphics[width=25px, height=20px]{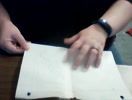}
		\includegraphics[width=25px, height=20px]{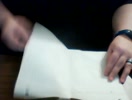}
		\includegraphics[width=25px, height=20px]{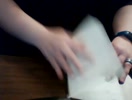}
		\includegraphics[width=25px, height=20px]{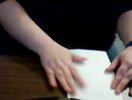} \vspace{-2mm}
		\caption{Folding something}
	\end{subfigure} \hfill
	\begin{subfigure}[b]{0.49\linewidth}\centering
		\captionsetup{labelformat=empty}
		\includegraphics[width=25px, height=20px]{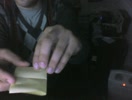}
		\includegraphics[width=25px, height=20px]{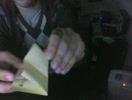}
		\includegraphics[width=25px, height=20px]{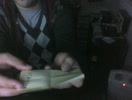}
		\includegraphics[width=25px, height=20px]{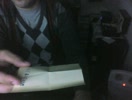}\vspace{-2mm}
		\caption{Unfolding something}
	\end{subfigure} \\
	\begin{subfigure}[b]{0.49\linewidth}\centering
		\captionsetup{labelformat=empty}
		\includegraphics[width=25px, height=20px]{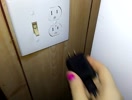}
		\includegraphics[width=25px, height=20px]{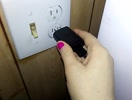}
		\includegraphics[width=25px, height=20px]{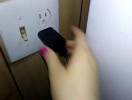}
		\includegraphics[width=25px, height=20px]{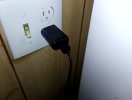}\vspace{-2mm}
		\caption{Plugging something into something}
	\end{subfigure} \hfill
	\begin{subfigure}[b]{0.49\linewidth}\centering
		\captionsetup{labelformat=empty}
		\includegraphics[width=25px, height=20px]{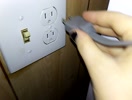}
		\includegraphics[width=25px, height=20px]{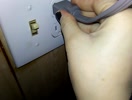}
		\includegraphics[width=25px, height=20px]{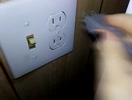}
		\includegraphics[width=25px, height=20px]{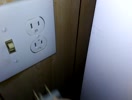}\vspace{-2mm}
		\caption{Plugging something into something but pulling it ...}
	\end{subfigure}\\    
	\begin{subfigure}[b]{0.49\linewidth}\centering
		\captionsetup{labelformat=empty}
		\includegraphics[width=25px, height=20px]{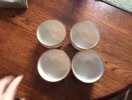}
		\includegraphics[width=25px, height=20px]{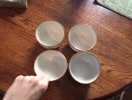}
		\includegraphics[width=25px, height=20px]{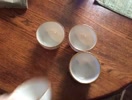}
		\includegraphics[width=25px, height=20px]{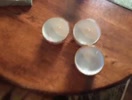}\vspace{-2mm}
		\caption{Taking one of many similar things on the table}
	\end{subfigure} \hfill
	\begin{subfigure}[b]{0.49\linewidth}\centering
		\captionsetup{labelformat=empty}
		\includegraphics[width=25px, height=20px]{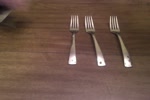}
		\includegraphics[width=25px, height=20px]{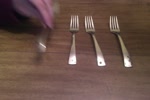}
		\includegraphics[width=25px, height=20px]{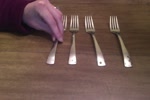}
		\includegraphics[width=25px, height=20px]{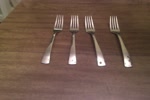}\vspace{-2mm}
		\caption{Putting something similar to other things that are ...}
	\end{subfigure} \\
	\begin{subfigure}[b]{0.49\linewidth}\centering
		\captionsetup{labelformat=empty}
		\includegraphics[width=25px, height=20px]{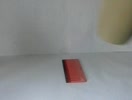}
		\includegraphics[width=25px, height=20px]{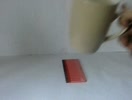}
		\includegraphics[width=25px, height=20px]{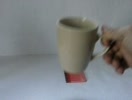}
		\includegraphics[width=25px, height=20px]{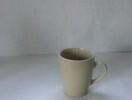}\vspace{-2mm}
		\caption{Putting something infront of something}
	\end{subfigure} \hfill
	\begin{subfigure}[b]{0.49\linewidth}\centering
		\captionsetup{labelformat=empty}
		\includegraphics[width=25px, height=20px]{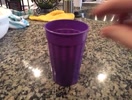}
		\includegraphics[width=25px, height=20px]{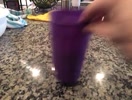}
		\includegraphics[width=25px, height=20px]{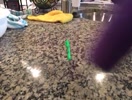}
		\includegraphics[width=25px, height=20px]{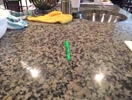}\vspace{-2mm}
		\caption{Removing something, revealing something behind}
	\end{subfigure}
	\captionsetup{font=normalsize, labelfont=normalsize}
	\caption{Sample frames from Something-V1 videos that belong to the most improved classes when \ac{gsm} is added.}
	\label{fig:something_samples}
\end{figure}
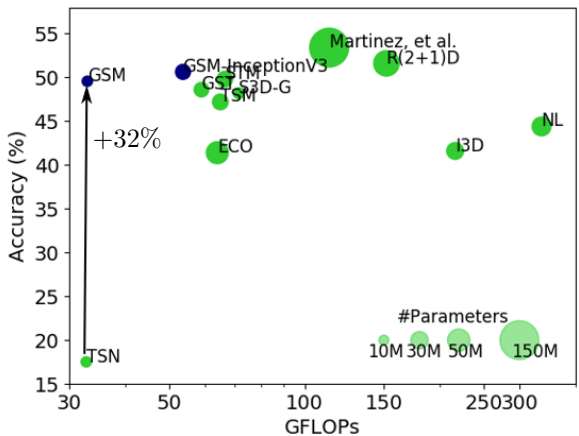
\begin{figure}
    \includegraphics[scale=0.4]{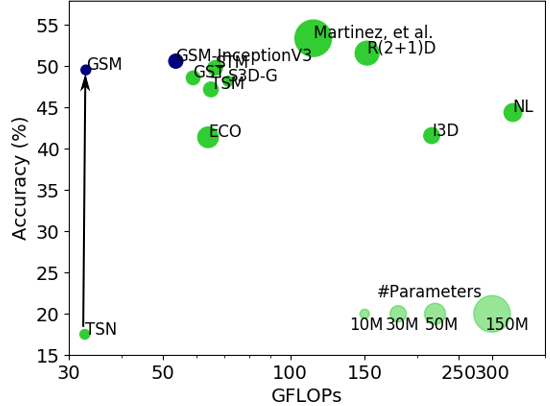}
    \begin{picture}(0,0)
    \put(-187,110){$+32\%$}
    \end{picture}
    \caption{Accuracy-vs-complexity of state-of-the-art on Something-V1, from Tab.~\ref{tab:something_sota}. Size indicates number of parameters (M, in millions). GSM outperforms or competes in recognition performance with far less model complexity.}
    \label{fig:complexity}
\end{figure}

In Fig.~\ref{fig:classes_imp}, we show the top 10 action classes that improved the most by adding \ac{gsm} to the CNN backbone of \ac{tsn}. From the figure, it can be seen that the network has enhanced its ability to distinguish between action classes that are similar in appearance, such as \verb+Unfolding something+ and \verb+Folding something+, \verb+Putting something infront of something+ and \verb+Removing something, revealing+ \verb+something behind+, etc. Sample frames from some of these most improved classes are shown in Fig.~\ref{fig:something_samples}. From these frames, we can see that reversing the order of the frames changes the action and thus the orderless pooling present in \ac{tsn} fails to identify the action. On the other hand, \ac{gsm} is able to improve the recognition score on these classes, providing a strong spatio-temporal reasoning. In order to validate the temporal encoding capability of \ac{gsm}, we evaluated its performance by applying the video frames in the reverse order. This resulted in a drastic degradation in the recognition performance from $47.24$ to $15.38\%$. On the other hand, there was no change in the recognition performance of \ac{tsn} when frames reversed in time were applied.

The t-SNE plots of the features from the final layer of the CNN corresponding to the 10 action groups described in~\cite{goyal2017something} are shown in Fig.~\ref{fig:tsn_tsne} and~\ref{fig:hftsn_tsne}. Fig.~\ref{fig:hftsn_tsne_improved} shows the t-SNE visualization of the most improved classes compared to \ac{tsn}. We sample 1800 videos from the validation split for the t-SNE visualization. It can be seen that the features from \ac{gsm} show a lower intra-class and higher inter-class variability compared to those from \ac{tsn}.

We also analyzed the memory requirement and computational complexity of \ac{gsm} and various \sota approaches. Fig.~\ref{fig:complexity} shows the accuracy, parameter and complexity trade-off computed on the validation set of Somthing-V1 dataset. The graph plots accuracy vs GFLOPs and the area of the bubbles indicate the number of parameters present in each method. From the plot, it can be seen that \ac{gsm} performs competitively to the \sota~\cite{martinez2019action} with less than one tenth the number of parameters and half the number of \acp{flop}.

\begin{table}\small
	\centering\renewcommand{\arraystretch}{0.7}
	\begin{tabular}{|c|c|c|}
		\hline
		\textbf{Method} & \textbf{Pre-training} & \textbf{Accuracy (\%)} \\ \hline \hline
		\ac{tsn}~\cite{tsn} (from~\cite{diving}) & ImageNet & 16.77 \\ \hline
		\acs{trn}~\cite{trn} (from~\cite{kanojia2019attentive}) & ImageNet & 22.8 \\ \hline
		R(2+1)D~\cite{r2plus1d} (from~\cite{dimofs}) & Kinetics & 28.9 \\ \hline
		DiMoFs~\cite{dimofs} & Kinetics & 31.4 \\ \hline
		P3D~\cite{p3d} (from~\cite{gst}) & ImageNet & 32.4 \\ \hline
		C3D~\cite{tran2015learning} (from~\cite{gst}) & ImageNet & 34.5 \\ \hline
		Kanojia~\etal~\cite{kanojia2019attentive} & ImageNet & 35.64 \\ \hline
		CorrNet~\cite{corrnet} & - & 37.7 \\ \hline
		\acs{gst} & ImageNet & 38.8  \\ \hline \hline 
		\ac{gsm} & ImageNet & 40.27 \\ \hline
	\end{tabular}
	\caption{Comparison to \sota on Diving48.}
	\label{tab:diving_sota}
\end{table}

\begin{table}[t]\small
	\centering \renewcommand{\arraystretch}{0.7}\tabcolsep=0.04cm 
	\begin{tabular}{|c|c|c|c|c|c|c|c|}
		\hline
		\multirow{2}{*}{\textbf{Method}} & \multirow{2}{*}{\textbf{Pre-train}} & \multicolumn{3}{c|}{\textbf{S1}} & \multicolumn{3}{c|}{\textbf{S2}}\\ \cline{3-8}
		& & Verb & Noun & Action & Verb & Noun & Action \\ \hline \hline
		\ac{tsn}~\cite{tsn} & ImageNet & 45.68 & 36.8 & 19.86 & 34.89 & 21.82 & 10.11 \\ \hline
		TBN~\cite{tbn} & ImageNet & 60.87 & 42.93 & 30.31 & 49.61 & 25.68 & 16.80 \\ \hline
		\acs{lsta}~\cite{lsta} & ImageNet & 59.55 & 38.35 & 30.33 & 47.32 & 22.16 & 16.63 \\ \hline
		RU-LSTM~\cite{rulstm} & ImageNet & 56.93 & 43.05 & 33.06 & 43.67 & 26.77 & 19.49 \\ \hline \hline
		LFB~\cite{lfb} & Kinetics & 60.0 & 45 & 32.7 & 50.9 & 31.5 & 21.2 \\ \hline
		R(2+1)D~\cite{ghadiyaram2019large} & Sports-1M & 59.6 & 43.7 & 31.0 & 47.2 & 28.7 & 18.3 \\ \hline
		R(2+1)D~\cite{ghadiyaram2019large} & External & 65.2 & 45.1 & 34.5 & 58.4 & 36.9 & 26.1 \\ \hline \hline
		\ac{gsm}& ImageNet & 59.41 & 41.83 & 33.45 & 48.28 & 26.15 & 20.18 \\ \hline
	\end{tabular}
	\caption{Comparison to \sota on EPIC-Kitchens.}
	\label{tab:epic_sota}
\end{table}

%% file: sub/5_supp.tex
This section includes details about the backbone CNN architecture, additional t-SNE plots,  results on Something Something-V1 dataset using ensemble of models, and visualization samples using saliency tubes. We also provide a supplementary video with the visualization samples. Code and models are available at \url{https://github.com/swathikirans/GSM}. 

\section{Architecture Details}
We provide the details of the CNN architectures used in our GSM models.

\subsection{BN-Inception}

Tab.~\ref{tab:architecture_BNInception} shows the architecture of GSM BN-Inception. The Inception modules used are shown in Fig.~4 of the paper. The table also lists the output size after each layer.

\begin{table}[th] \centering\scriptsize
	\begin{tabular}{|c|c|c|}
		\hline
		\multirow{2}{*}{Type} & Kernel size/ & \multirow{2}{*}{Output size}\\
		& stride & \\  
		\hline \hline
		Conv & $7\times7/2$ & $112\times112\times64$\\ \hline
		Max Pool & $3\times3/2$ & $56\times56\times64$\\ \hline
		Conv & $1\times1/1$ & $56\times56\times64$\\ \hline
		Conv & $3\times3/1$ & $56\times56\times192$\\ \hline
		Max Pool & $3\times3/2$ & $28\times28\times192$\\ \hline
		Inception-GSM 1 (Inc3a) &  & $28\times28\times256$\\ \hline
		Inception-GSM 1 (Inc3b) &  & $28\times28\times320$\\ \hline
		Inception-GSM 2 (Inc3c) &  & $14\times14\times576$\\ \hline
		Inception-GSM 1 (Inc4a) &  & $14\times14\times576$\\ \hline
		Inception-GSM 1 (Inc4b) &  & $14\times14\times576$\\ \hline
		Inception-GSM 1 (Inc4c) &  & $14\times14\times608$\\ \hline
		Inception-GSM 1 (Inc4d) &  & $14\times14\times608$\\ \hline
		Inception-GSM 2 (Inc4e) &  & $7\times7\times1056$\\ \hline
		Inception-GSM 1 (Inc5a) &  & $7\times7\times1024$\\ \hline
		Inception-GSM 1 (Inc5b) &  & $7\times7\times1024$\\ \hline
		Avg Pool & $7\times7/1$ &$ 1\times1\times1024$\\ \hline
		Linear & & $1\times1\times C$ \\ \hline
	\end{tabular}
	\caption{Gate-Shift BN-Inception Architecture. All convolution layers are followed by \ac{bn} layer and ReLU non-linearity. $C$ is the number of classes in the dataset.}
	\label{tab:architecture_BNInception}
\end{table}

\subsection{InceptionV3}
\begin{figure*}
	\centering
	\begin{subfigure}[b]{0.19\linewidth}
		\centering
		\includegraphics[scale=.075]{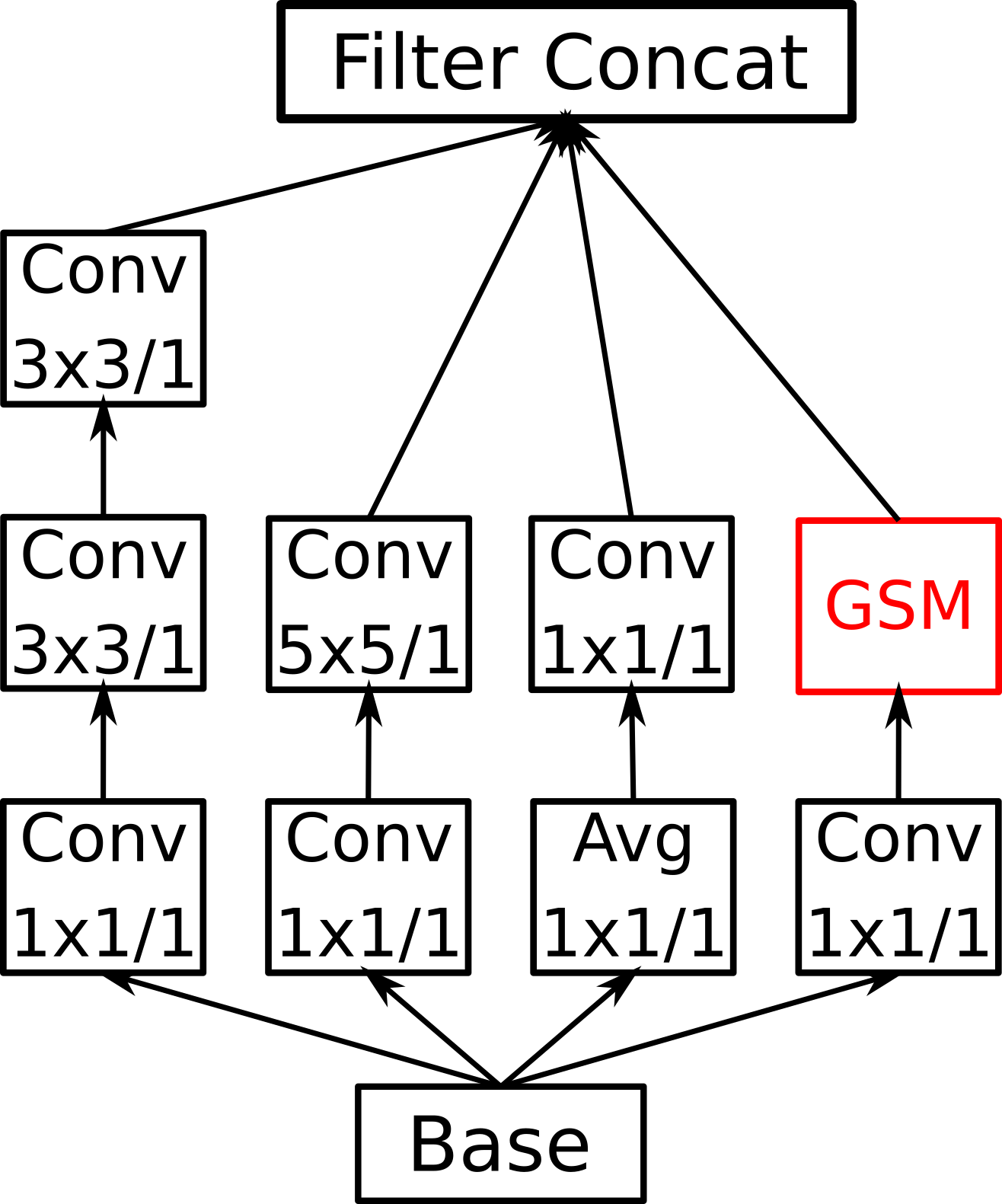}
		\caption{}
	\end{subfigure} \hfill
	\begin{subfigure}[b]{0.19\linewidth}
		\centering
		\includegraphics[scale=.075]{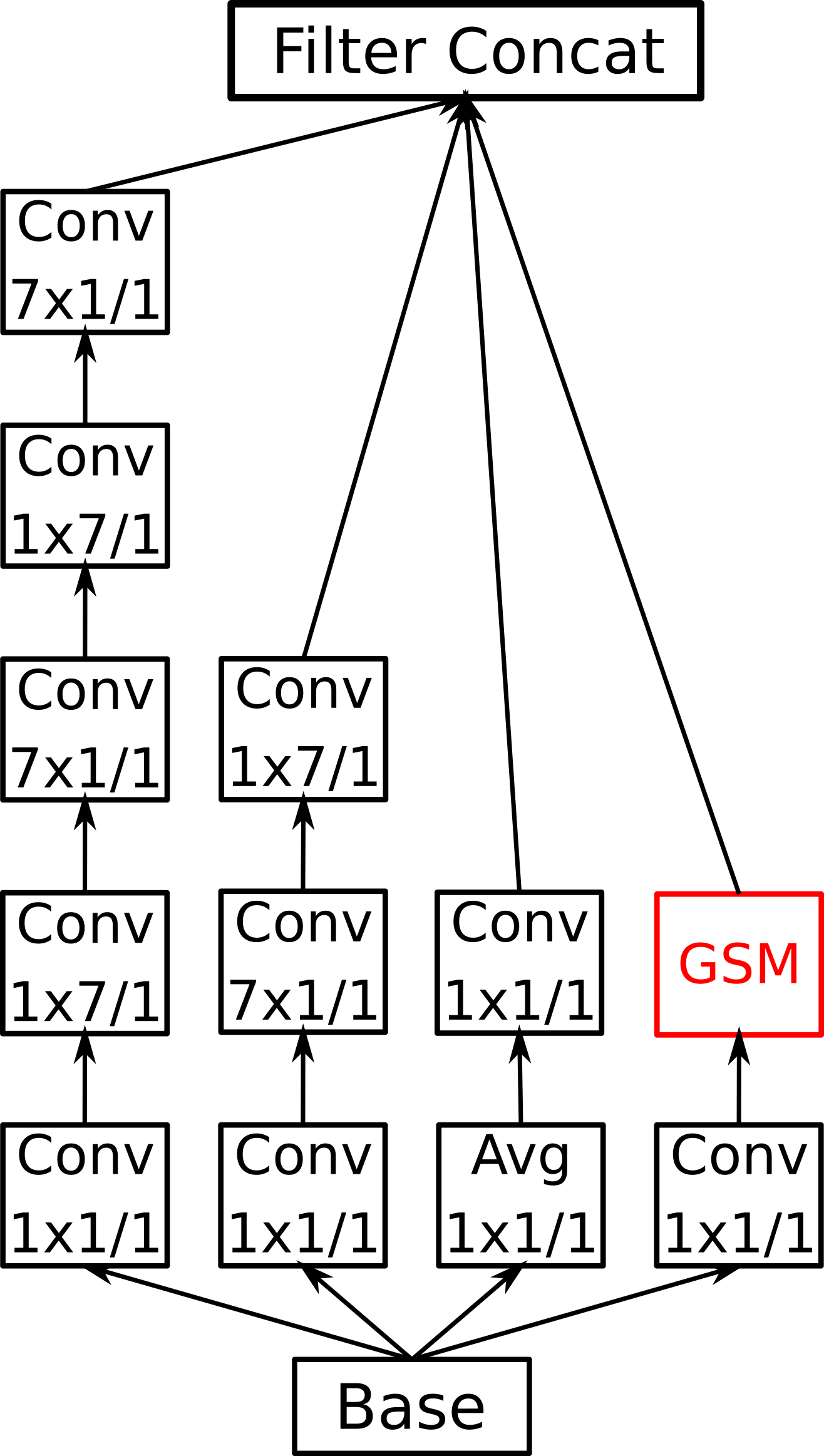}
		\caption{}
	\end{subfigure}\hfill
	\begin{subfigure}[b]{0.16\linewidth}
		\centering
		\includegraphics[scale=.06]{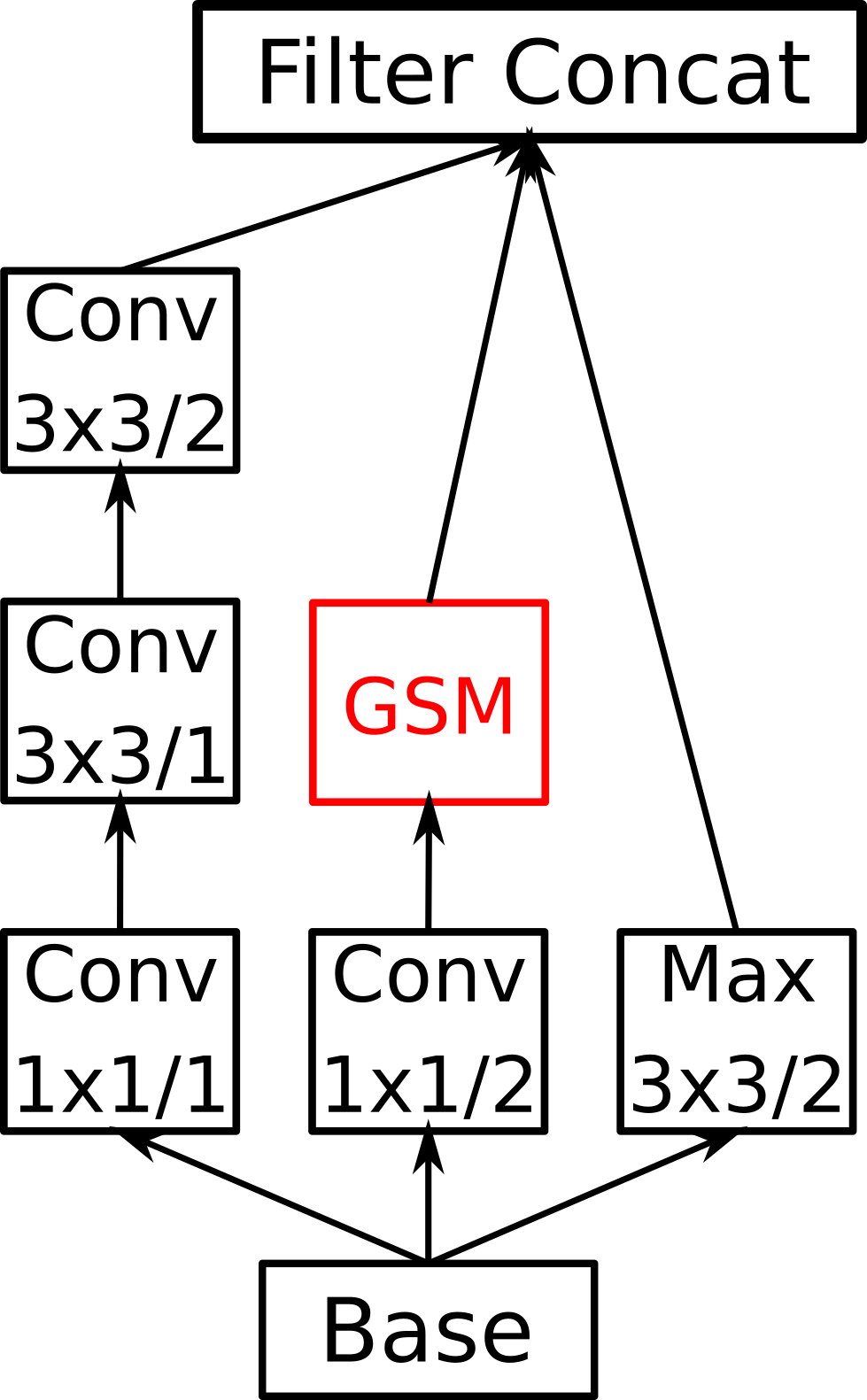}
		\caption{}
	\end{subfigure} \hfill
	\begin{subfigure}[b]{0.16\linewidth}
		\centering
		\includegraphics[scale=.06]{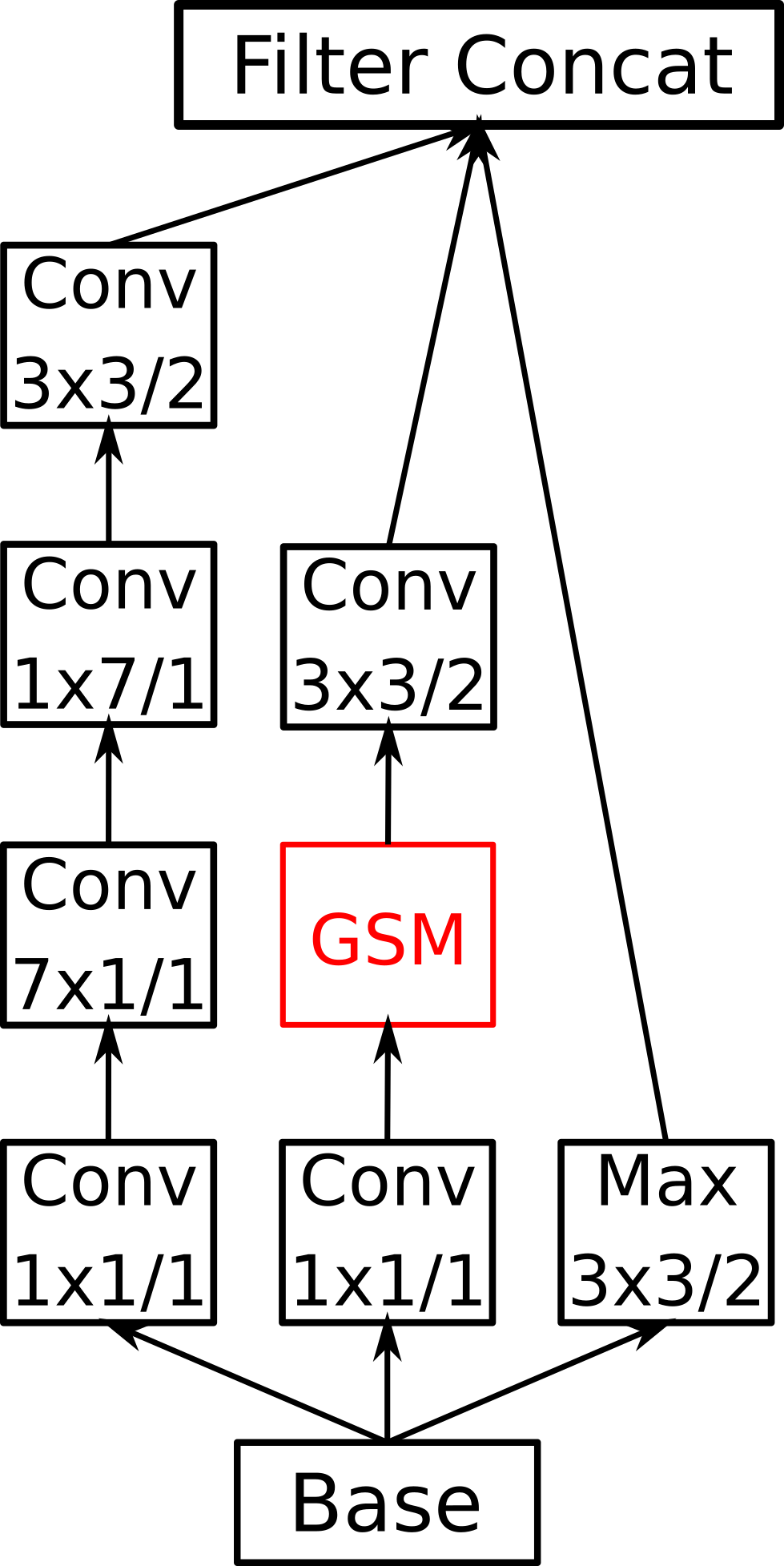}
		\caption{}
	\end{subfigure} \hfill
	\begin{subfigure}[b]{0.21\linewidth}
		\centering
		\includegraphics[scale=.06]{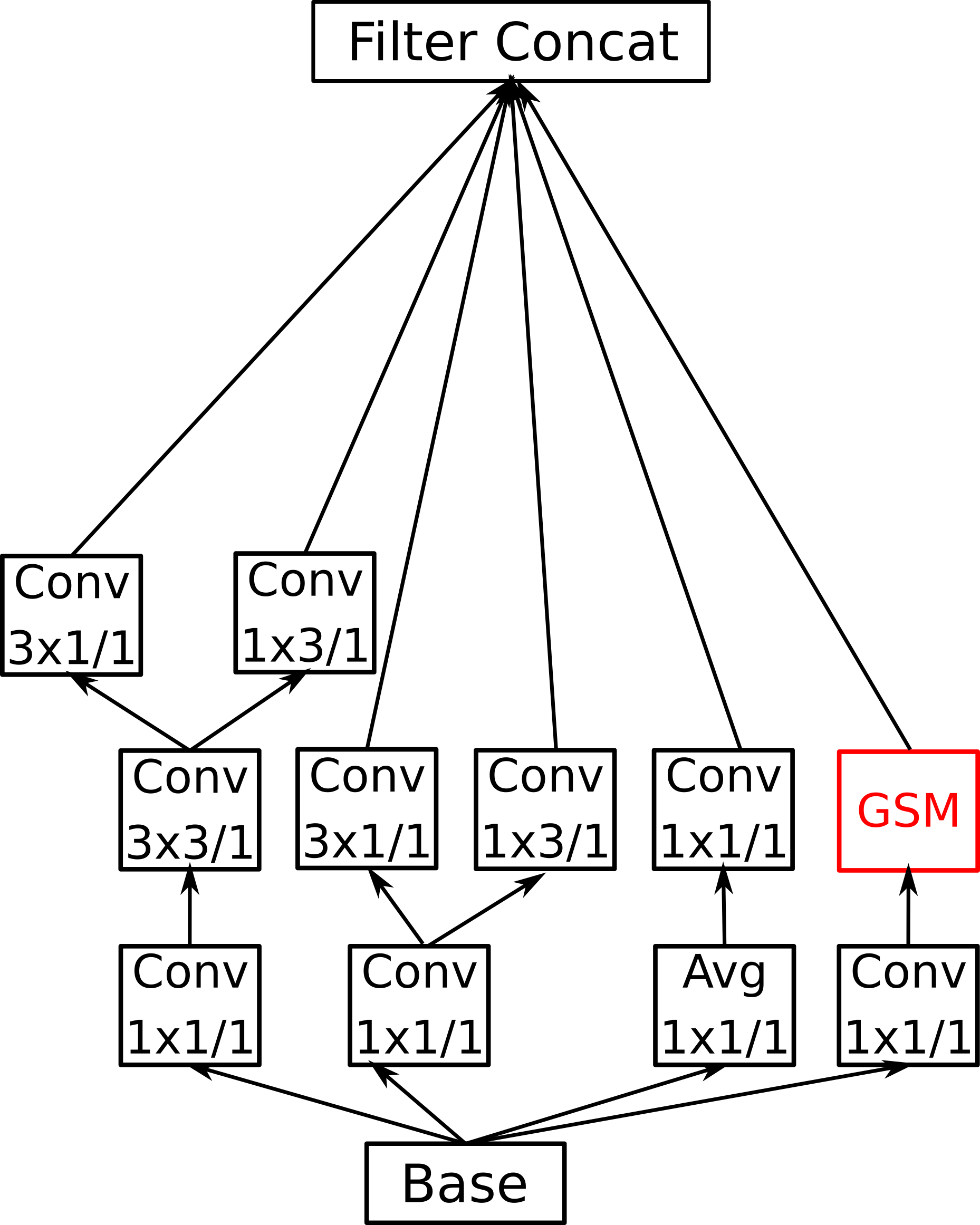}
		\caption{}
	\end{subfigure} \vskip -2mm
	\caption{Inception blocks with GSM used in the InceptionV3 architecture.}
	\label{fig:inceptionV3}
\end{figure*}

The architecture of GSM InceptionV3 is shown in Tab.~\ref{tab:architecture_inceptionv3} along with the size of the outputs after each layer. We apply an input of size $229\times229$ instead of the standard size of $299\times299$. This reduces the computational complexity without affecting the performance of the model. The Inception blocks with GSM used in the model are shown in Fig.~\ref{fig:inceptionV3}.

\begin{table}[th] \centering\scriptsize
	\begin{tabular}{|c|c|c|}
		\hline
		\multirow{2}{*}{Type} & Kernel size/ & \multirow{2}{*}{Output size}\\
		& stride & \\  
		\hline \hline
		Conv & $3\times3/2$ & $114\times114\times32$\\ \hline
		Conv & $3\times3/1$ & $112\times112\times32$\\ \hline
		Conv & $3\times3/1$ & $112\times112\times64$\\ \hline
		Max Pool & $3\times3/2$ & $56\times56\times64$\\ \hline
		Conv & $3\times3/1$ & $56\times56\times80$\\ \hline
		Conv & $3\times3/1$ & $54\times54\times192$\\ \hline
		Max Pool & $3\times3/2$ & $27\times27\times192$\\ \hline
		Inception-GSM (Fig. 1a) &  & $27\times27\times256$\\ \hline
		Inception-GSM (Fig. 1a) &  & $27\times27\times288$\\ \hline
		Inception-GSM (Fig. 1c) &  & $27\times27\times288$\\ \hline
		Inception-GSM (Fig. 1b) &  & $13\times13\times768$\\ \hline
		Inception-GSM (Fig. 1b) &  & $13\times13\times768$\\ \hline
		Inception-GSM (Fig. 1b) &  & $13\times13\times768$\\ \hline
		Inception-GSM (Fig. 1b) &  & $13\times13\times768$\\ \hline
		Inception-GSM (Fig. 1b) &  & $13\times13\times768$\\ \hline
		Inception-GSM (Fig. 1d) &  & $6\times6\times1280$\\ \hline
		Inception-GSM (Fig. 1e) &  & $6\times6\times2048$\\ \hline
		Inception-GSM (Fig. 1e) &  & $6\times6\times2048$\\ \hline
		Avg Pool & $6\times6/1$ &$ 1\times1\times2048$\\ \hline
		Linear & & $1\times1\times C$ \\ \hline
	\end{tabular}
	\caption{Gate-Shift InceptionV3 Architecture. All convolution layers are followed by \ac{bn} layer and ReLU non-linearity. $C$ is the number of classes in the dataset.}
	\label{tab:architecture_inceptionv3}
\end{table}

\section{t-SNE }

\begin{figure*}
    \centering
    \begin{subfigure}[b]{0.45\linewidth}
    \centering
        \includegraphics[width=180px]{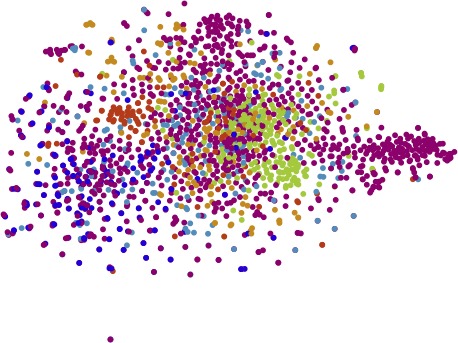} \vskip -2mm
        \caption{No GSM}
        \label{fig:no_gsm}
    \end{subfigure} \hfill
    \begin{subfigure}[b]{0.45\linewidth}
        \centering
        \includegraphics[width=180px]{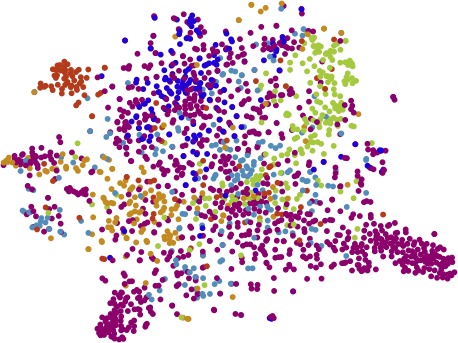} \vskip -2mm
        \caption{1 GSM}
        \label{fig:1_gsm}
    \end{subfigure}\\[.2in]
        \begin{subfigure}[b]{0.45\linewidth}
    \centering
        \includegraphics[width=180px]{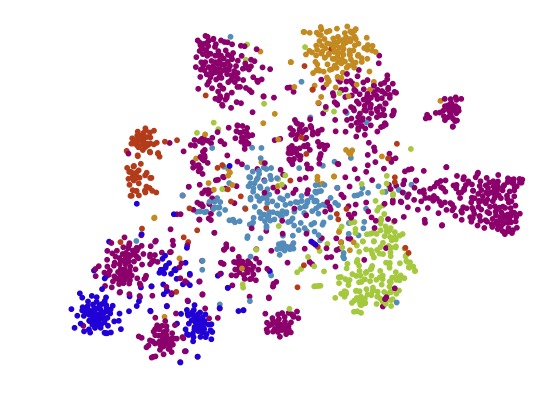}
        \caption{5 GSM}
        \label{fig:5_gsm}
    \end{subfigure} \hfill
    \begin{subfigure}[b]{0.45\linewidth}
        \centering
        \includegraphics[width=180px]{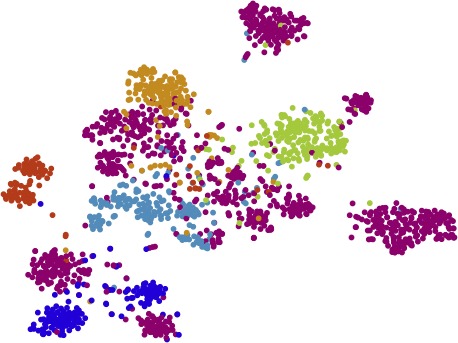}
        \caption{10 GSM}
        \label{fig:10_gsm}
    \end{subfigure}
    \vspace{-0.2cm}
    \caption{t-SNE visualization of features from networks that use (a) No GSM, (b) 1 GSM, (c) 5 GSMs and (d) 10 GSMs.}
    \label{fig:tsne_ablation}
\end{figure*}

\begin{figure*}
	\centering

	\begin{subfigure}[b]{0.45\linewidth}
		\centering
		\includegraphics[width=180px]{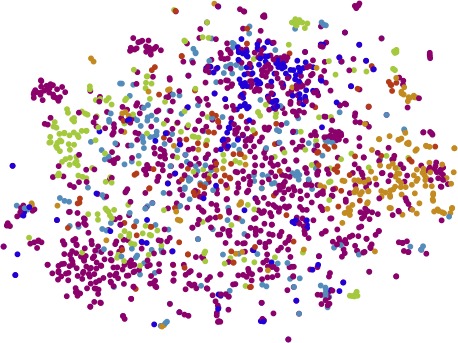}
		\caption{Inc4d}
	\end{subfigure}\hfill
	\begin{subfigure}[b]{0.45\linewidth}
		\centering
		\includegraphics[width=180px]{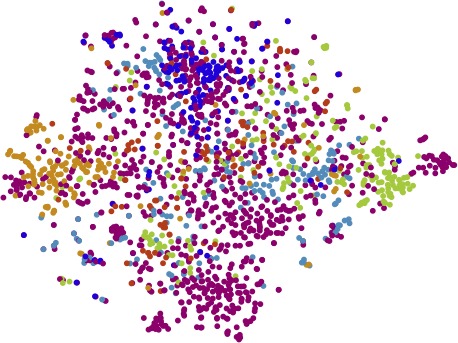}
		\caption{Inc4e}
	\end{subfigure} \\[.2in]
	\begin{subfigure}[b]{0.45\linewidth}
		\centering
		\includegraphics[width=180px]{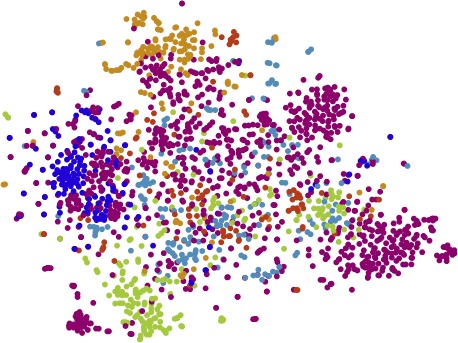}
		\caption{Inc5a}
	\end{subfigure}\hfill
	\begin{subfigure}[b]{0.45\linewidth}
		\centering
		\includegraphics[width=180px]{./figures/tsne/gsm_feats_feats_gsm_BN16_10groups_new_cropped.jpg}
		\caption{Inc5b}
	\end{subfigure}
	\caption{t-SNE visualization of features from intermediate layer of GSM BN-Inception.}
	\label{fig:tsne_gsm}
\end{figure*}

We first visualize the t-SNE plot of features for the models used in the ablation study, \ie, model with no GSM (Fig.~\ref{fig:no_gsm}), model with 1 GSM (Fig.~\ref{fig:1_gsm}), model with 5 GSM (Fig.~\ref{fig:5_gsm}) and model with 10 GSM (Fig.~\ref{fig:10_gsm}). All figures plot the features of the 10 action groups presented in \cite{goyal2017something}. From the figures, one can see that adding GSM into the CNN results in a reduction of intra-class variability and in an increase of inter-class variability. 
Fig.~\ref{fig:tsne_gsm} shows the t-SNE plot of features from the last four Inception blocks of BN-Inception with 10 GSM. From the figure, we can see that the semantic separation increases as we move towards the top layers of the backbone.

\section{Ensemble Results}
\begin{figure*}[t]\centering
\includegraphics[scale=0.5]{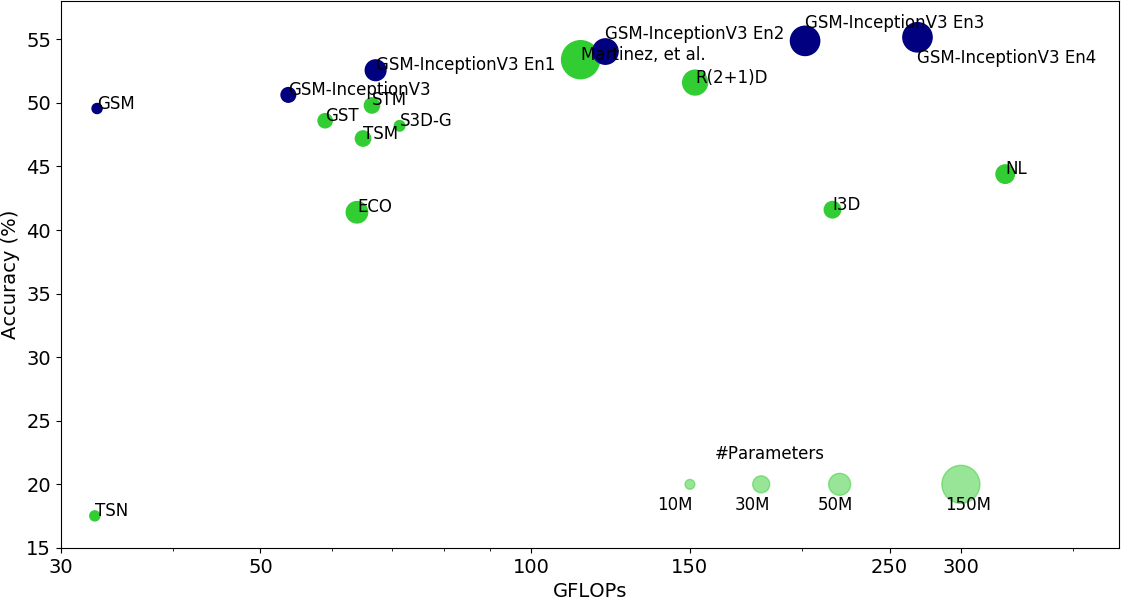}
\vspace{-0.3cm}
\caption{Accuracy-vs-complexity of state-of-the-art on Something-V1. Size indicates number of parameters (M, in millions).}
\label{fig:complexity_ensemble}
\end{figure*}

	Tab.~\ref{tab:ensemble} lists the action recognition accuracy, number of parameters and FLOPS obtained by ensembling the models presented in this work on Something Something-V1 dataset. The first and second blocks in the table list the accuracy obtained with individual models when evaluated using 1 and 2 clips, respectively. The third block shows the recognition accuracy with different ensemble models. Ensembling is done by combining GSM InceptionV3 models that are trained with different number of input frames. We average the prediction scores obtained from individual models to compute the performance of the ensemble. From the table, it can be seen that the accuracy is increasing as more models are being added. Using models with different number of input frames enables the ensemble to encode the video with different temporal resolutions. Such an ensemble has some analogy with SlowFast~\cite{slowfast}. With an ensemble of models trained on 8, 12, 16 and 24 frames, we achieve a state-of-the-art recognition accuracy of 55.16\%. We include the parameter and complexity trade-off in Fig.~\ref{fig:complexity_ensemble}. From the figure, we can see that the ensemble of GSM family achieves the state-of-the-art recognition performance with fewer parameters than previous state-of-the-art~\cite{martinez2019action}.
	
	\begin{table*}[ht]
		\centering
		\begin{tabular}{|c|c|c|c|c|}
			\hline
			\textbf{Model} & \textbf{\#Frames} & \textbf{Params. (M)} & \textbf{FLOPs (G)} & \textbf{Accuracy (\%)} \\ \hline \hline
			GSM InceptionV3 & 8 & 22.21 & 26.85 & 49.01\\ \hline
			GSM InceptionV3 & 12 & 22.21 & 40.26 & 51.58\\ \hline
			GSM InceptionV3 & 16 & 22.21 & 53.7 & 50.63\\ \hline
			GSM InceptionV3 & 24 & 22.21 & 80.55 & 49.63\\ \hline \hline
			GSM InceptionV3 & 8$\times$2 & 22.21 & 53.7 & 50.43\\ \hline
			GSM InceptionV3 & 12$\times$2 & 22.21 & 80.55 & 51.98\\ \hline
			GSM InceptionV3 & 16$\times$2 & 22.21 & 107.4 & 51.68\\ \hline
			GSM InceptionV3 & 24$\times$2 & 22.21 & 161.1 & 50.35\\ \hline \hline
			GSM InceptionV3 En1 & 8+12 & 44.42 & 67.13 & 52.57\\ \hline
			GSM InceptionV3 En2 & 8+12+16 & 66.63 & 120.83 & 54.04\\ \hline
			GSM InceptionV3 En3 & 8+12+16+24 & 88.84 & 201.38 & 54.88\\ \hline
			GSM InceptionV3 En3 & 8$\times$2+12$\times$2+16+24 & 88.84 & 268.47 & 55.16\\ \hline
		\end{tabular}
		\caption{Recognition Accuracy obtained on Something Something-V1 dataset by ensembling different models.}
		\label{tab:ensemble}\vspace{-0.3cm}
	\end{table*}

\section{Visualization}

\begin{figure*}[t]
		\centering
		\begin{subfigure}[b]{0.005\textwidth} 
			\raisebox{4in}{\rotatebox[origin=c]{90}{\texttt{putting something in front of something}}}
		\end{subfigure} \hskip -7mm
		\begin{subfigure}[b]{0.24\textwidth}
			\centering
			\includegraphics[scale=0.15]{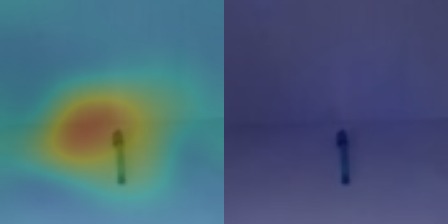} \\
			\includegraphics[scale=0.15]{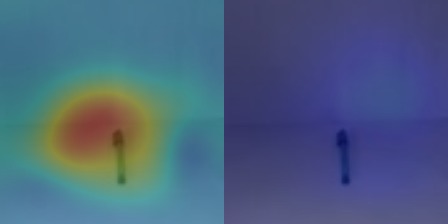} \\
			\includegraphics[scale=0.15]{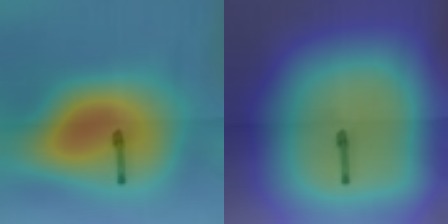} \\
			\includegraphics[scale=0.15]{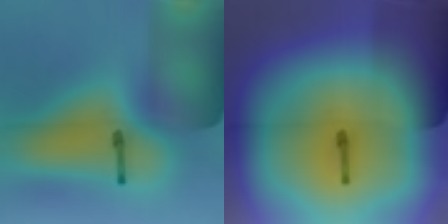} \\
			\includegraphics[scale=0.15]{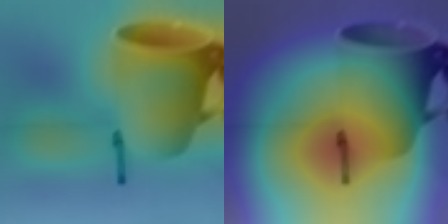} \\
			\includegraphics[scale=0.15]{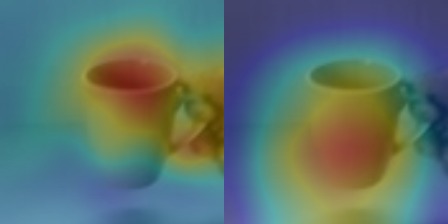} \\
			\includegraphics[scale=0.15]{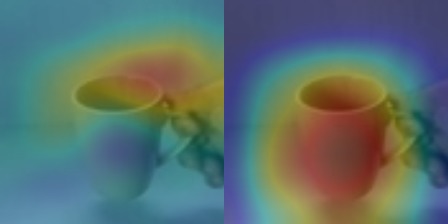} \\
			\includegraphics[scale=0.15]{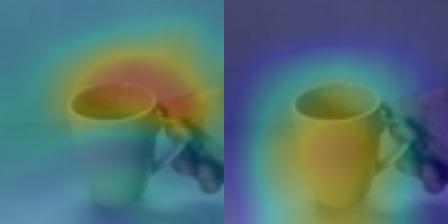} \\
			\includegraphics[scale=0.15]{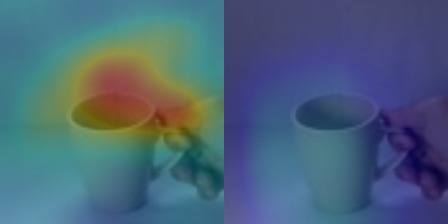} \\
			\includegraphics[scale=0.15]{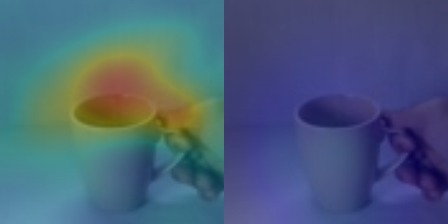} \\
			\includegraphics[scale=0.15]{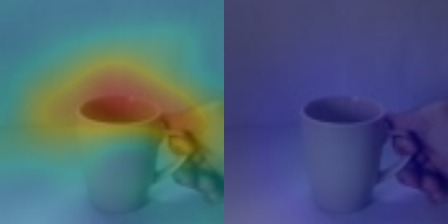} \\
			\includegraphics[scale=0.15]{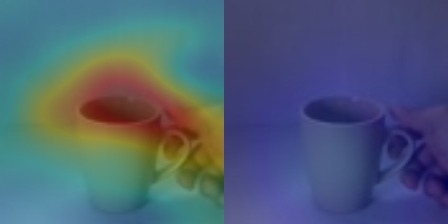} \\
			\includegraphics[scale=0.15]{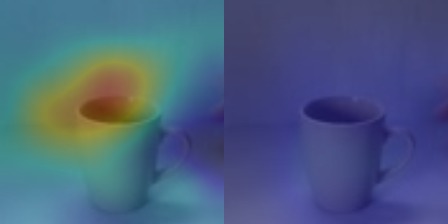} \\
			\includegraphics[scale=0.15]{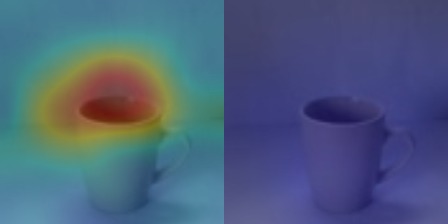} \\
			\includegraphics[scale=0.15]{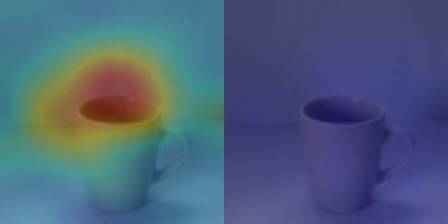} \\
			\includegraphics[scale=0.15]{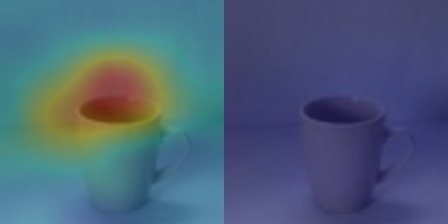} 
			\caption{}
			\label{fig:put_object}
		\end{subfigure}
		\begin{subfigure}[b]{0.005\textwidth} 
			\raisebox{4in}{\rotatebox[origin=c]{90}{\texttt{unfolding something}}}
		\end{subfigure} \hskip -7mm
		\begin{subfigure}[b]{0.24\textwidth}
			\centering
			\includegraphics[scale=0.15]{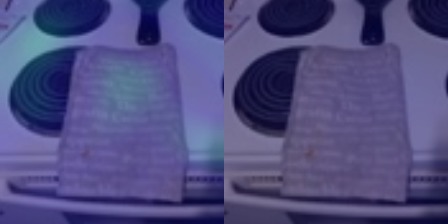} \\
			\includegraphics[scale=0.15]{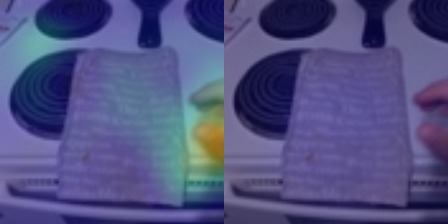} \\
			\includegraphics[scale=0.15]{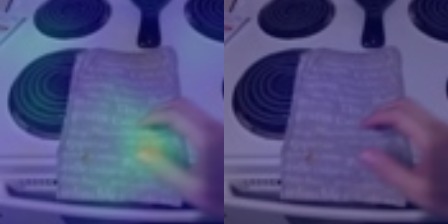} \\
			\includegraphics[scale=0.15]{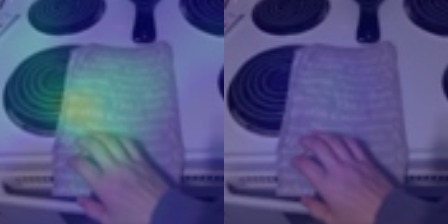} \\
			\includegraphics[scale=0.15]{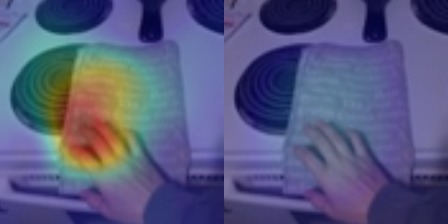} \\
			\includegraphics[scale=0.15]{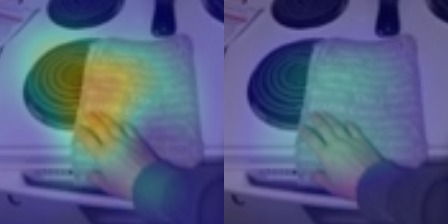} \\
			\includegraphics[scale=0.15]{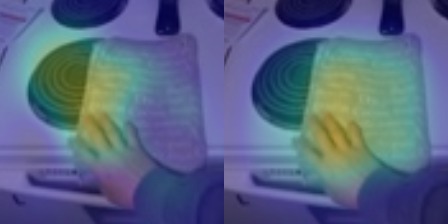} \\
			\includegraphics[scale=0.15]{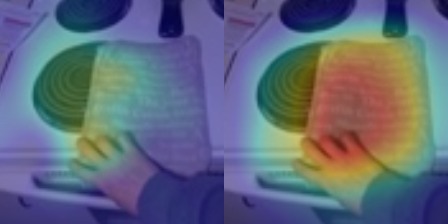} \\
			\includegraphics[scale=0.15]{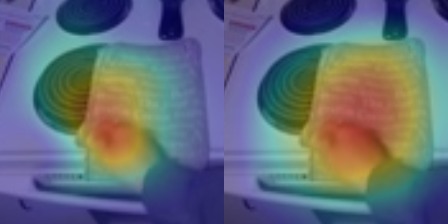} \\
			\includegraphics[scale=0.15]{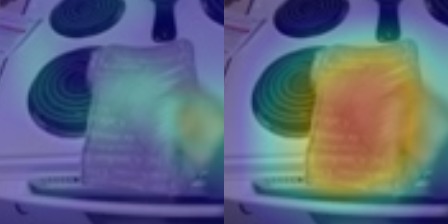} \\
			\includegraphics[scale=0.15]{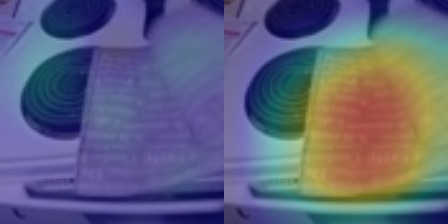} \\
			\includegraphics[scale=0.15]{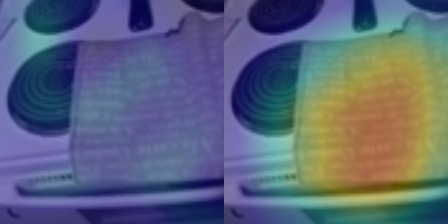} \\
			\includegraphics[scale=0.15]{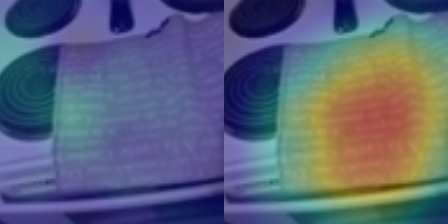} \\
			\includegraphics[scale=0.15]{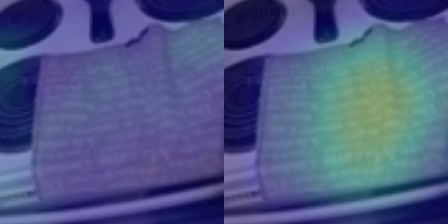} \\
			\includegraphics[scale=0.15]{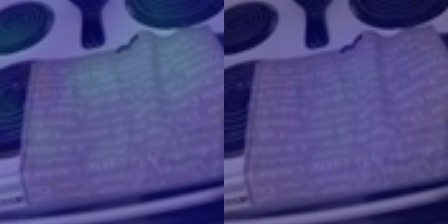} \\
			\includegraphics[scale=0.15]{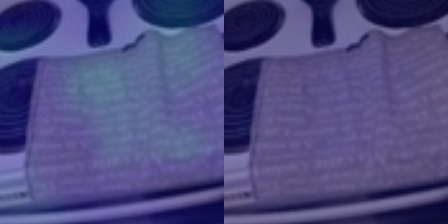}  
			\caption{}
		\end{subfigure} 
		\begin{subfigure}[b]{0.005\textwidth} 
			\raisebox{4in}{\rotatebox[origin=c]{90}{\texttt{removing something, revealing something behind}}}
		\end{subfigure} \hskip -7mm
		\begin{subfigure}[b]{0.24\textwidth}
			\centering
			\includegraphics[scale=0.15]{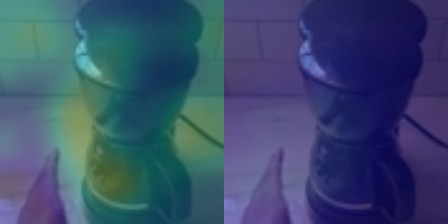} \\
			\includegraphics[scale=0.15]{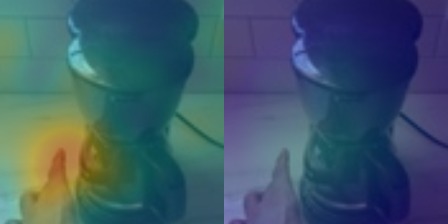} \\
			\includegraphics[scale=0.15]{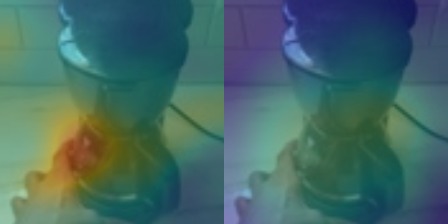} \\
			\includegraphics[scale=0.15]{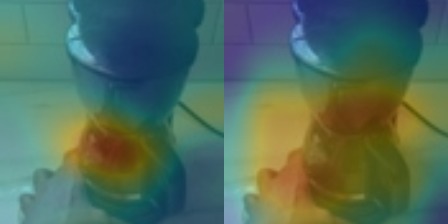} \\
			\includegraphics[scale=0.15]{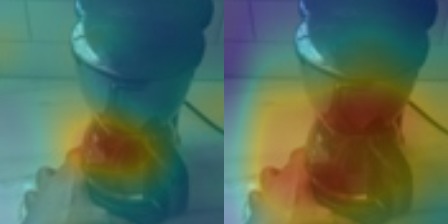} \\
			\includegraphics[scale=0.15]{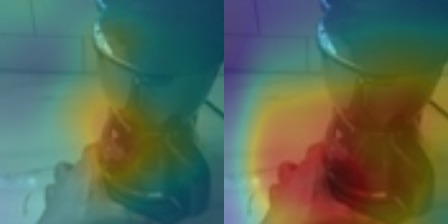} \\
			\includegraphics[scale=0.15]{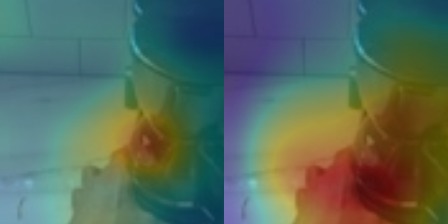} \\
			\includegraphics[scale=0.15]{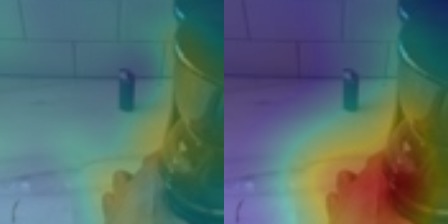} \\
			\includegraphics[scale=0.15]{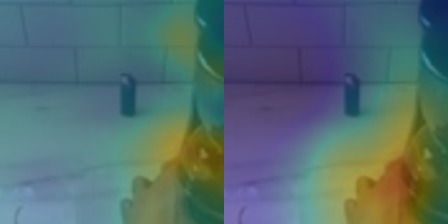} \\
			\includegraphics[scale=0.15]{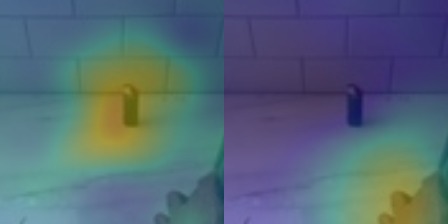} \\
			\includegraphics[scale=0.15]{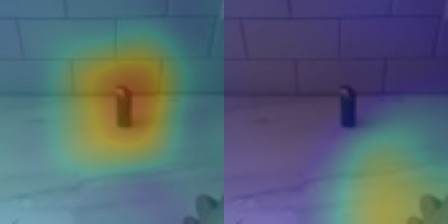} \\
			\includegraphics[scale=0.15]{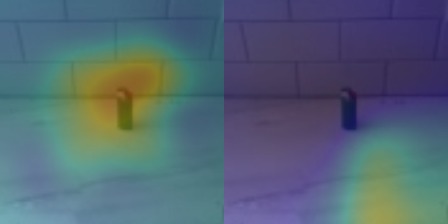} \\
			\includegraphics[scale=0.15]{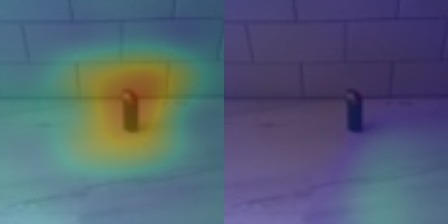} \\
			\includegraphics[scale=0.15]{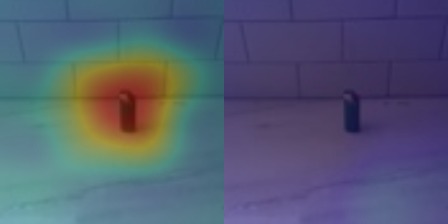} \\
			\includegraphics[scale=0.15]{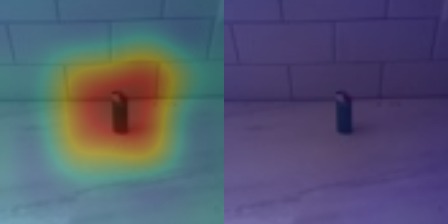} \\
			\includegraphics[scale=0.15]{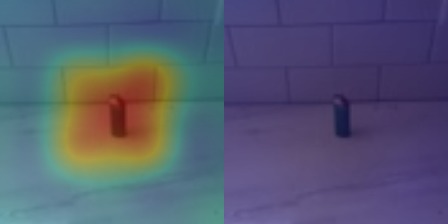} 
			\caption{} 
		\end{subfigure}
		\begin{subfigure}[b]{0.005\textwidth} 
			\raisebox{4in}{\rotatebox[origin=c]{90}{\texttt{taking one of many similar things on the table}}}
		\end{subfigure} \hskip -7mm
		\begin{subfigure}[b]{0.24\textwidth}
			\centering
			\includegraphics[scale=0.15]{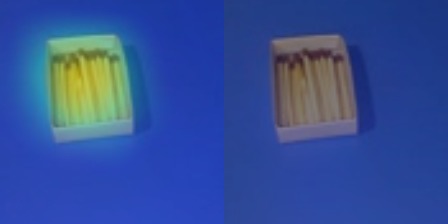} \\
			\includegraphics[scale=0.15]{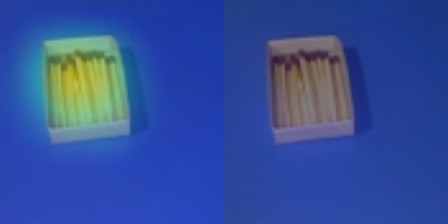} \\
			\includegraphics[scale=0.15]{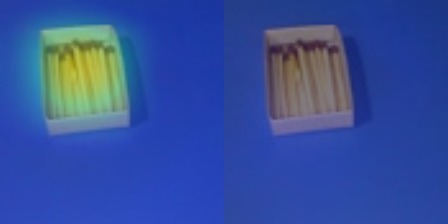} \\
			\includegraphics[scale=0.15]{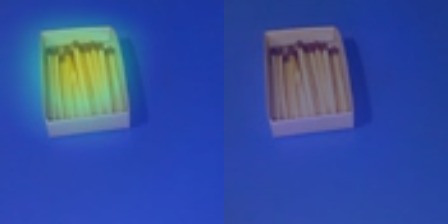} \\
			\includegraphics[scale=0.15]{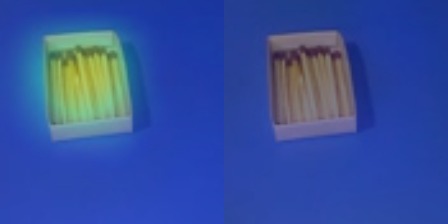} \\
			\includegraphics[scale=0.15]{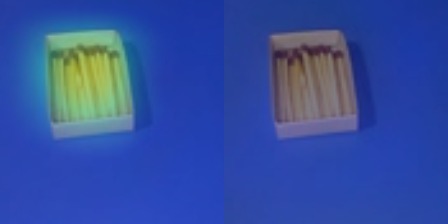} \\
			\includegraphics[scale=0.15]{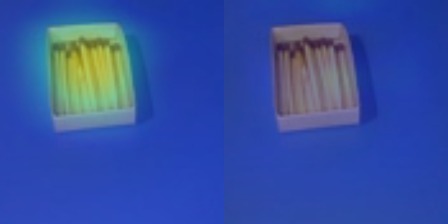} \\
			\includegraphics[scale=0.15]{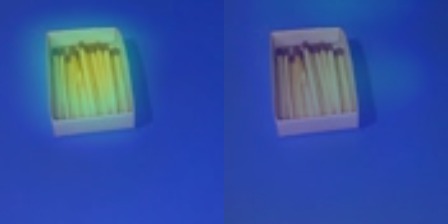} \\
			\includegraphics[scale=0.15]{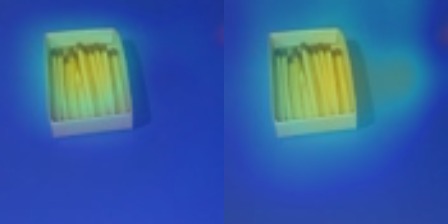} \\
			\includegraphics[scale=0.15]{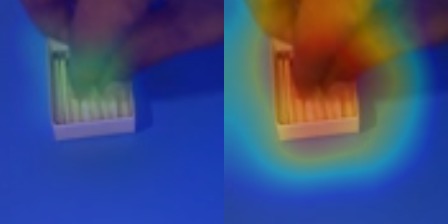} \\
			\includegraphics[scale=0.15]{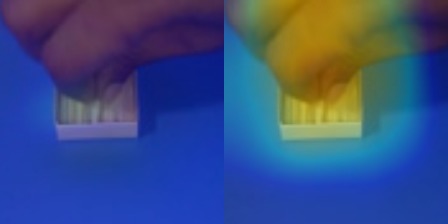} \\
			\includegraphics[scale=0.15]{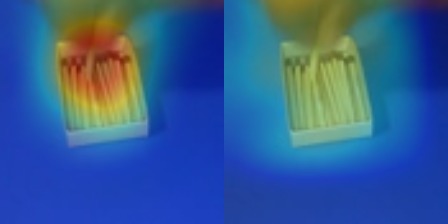} \\
			\includegraphics[scale=0.15]{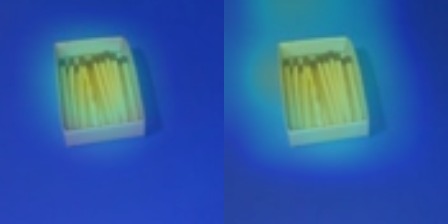} \\
			\includegraphics[scale=0.15]{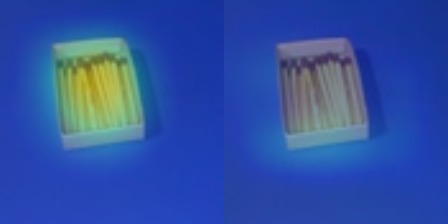} \\
			\includegraphics[scale=0.15]{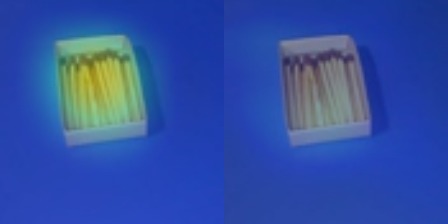} \\
			\includegraphics[scale=0.15]{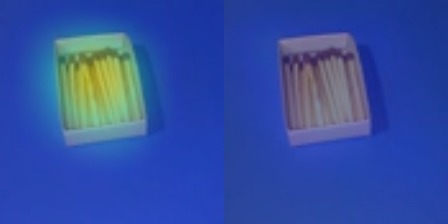}
			\caption{}
			\label{fig:remove}
		\end{subfigure}
		\caption{Saliency tubes generated by TSN (left) and GSM (right) on sample videos taken from the validation set of Something Something-V1 dataset. Action labels are shown as text on columns.
			}
		\label{fig:cam1}
	\end{figure*}

\begin{figure*}[t]
		\centering 
		\begin{subfigure}[b]{0.005\textwidth} 
		\raisebox{4in}{\rotatebox[origin=c]{90}{\texttt{plugging something into something}}}
		\end{subfigure} \hskip -7mm
		\begin{subfigure}[b]{0.24\textwidth}  
			\centering 
			\includegraphics[scale=0.15]{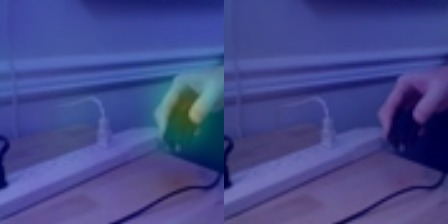} \\
			\includegraphics[scale=0.15]{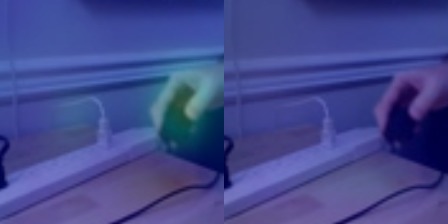} \\
			\includegraphics[scale=0.15]{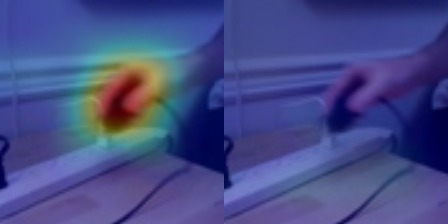} \\
			\includegraphics[scale=0.15]{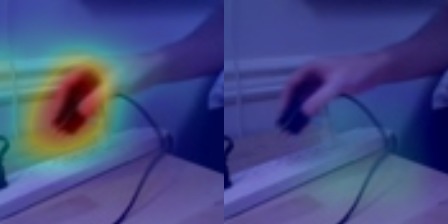} \\
			\includegraphics[scale=0.15]{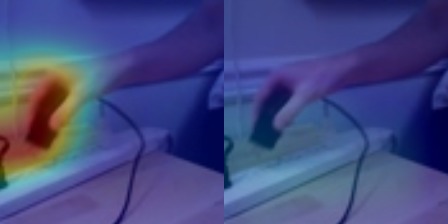} \\
			\includegraphics[scale=0.15]{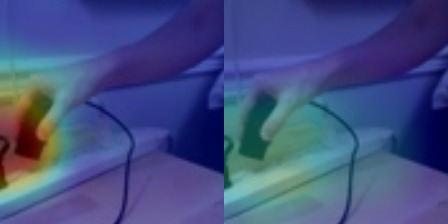} \\
			\includegraphics[scale=0.15]{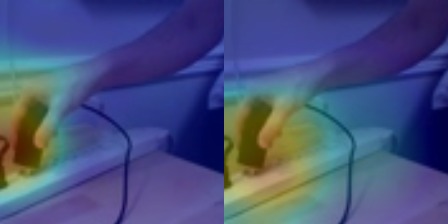} \\
			\includegraphics[scale=0.15]{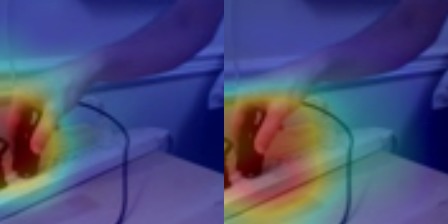} \\
			\includegraphics[scale=0.15]{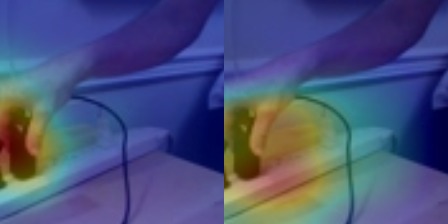} \\
			\includegraphics[scale=0.15]{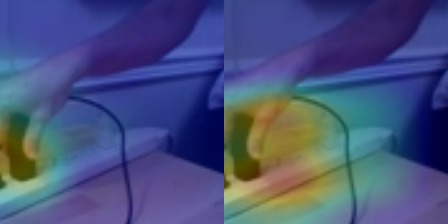} \\
			\includegraphics[scale=0.15]{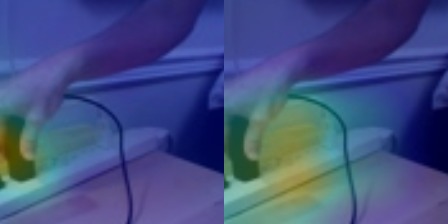} \\
			\includegraphics[scale=0.15]{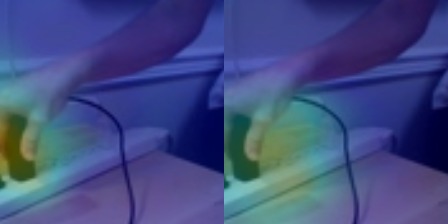} \\
			\includegraphics[scale=0.15]{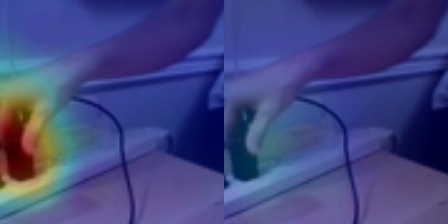} \\
			\includegraphics[scale=0.15]{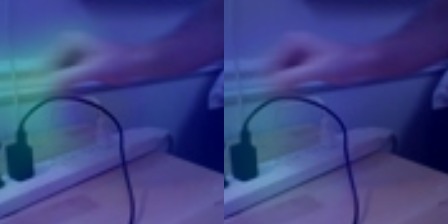} \\
			\includegraphics[scale=0.15]{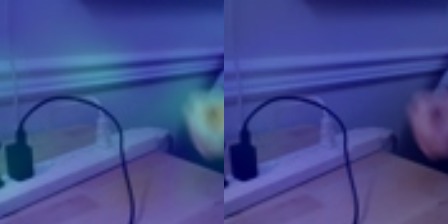} \\
			\includegraphics[scale=0.15]{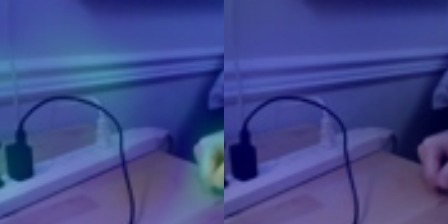}  
			\caption{}
		\end{subfigure}
		\begin{subfigure}[b]{0.005\textwidth} 
			\raisebox{4in}{\rotatebox[origin=c]{90}{\texttt{folding something}}}
		\end{subfigure} \hskip -7mm
		\begin{subfigure}[b]{0.24\textwidth}
			\centering
			\includegraphics[scale=0.15]{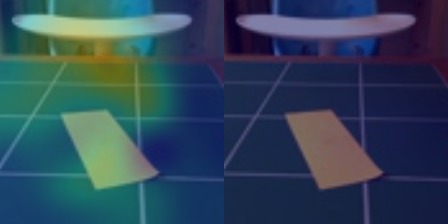} \\
			\includegraphics[scale=0.15]{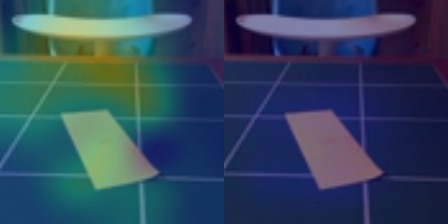} \\
			\includegraphics[scale=0.15]{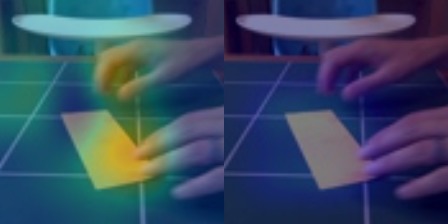} \\
			\includegraphics[scale=0.15]{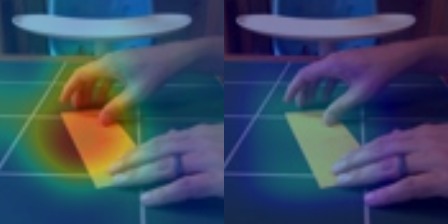} \\
			\includegraphics[scale=0.15]{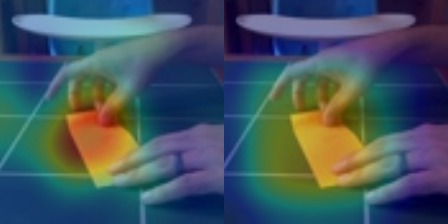} \\
			\includegraphics[scale=0.15]{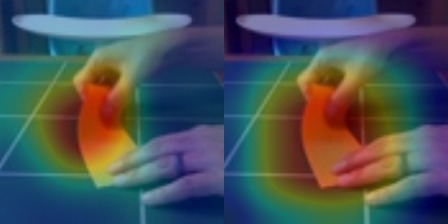} \\
			\includegraphics[scale=0.15]{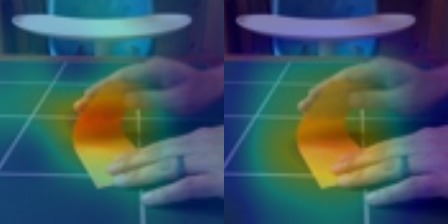} \\
			\includegraphics[scale=0.15]{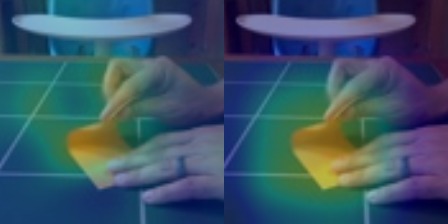} \\
			\includegraphics[scale=0.15]{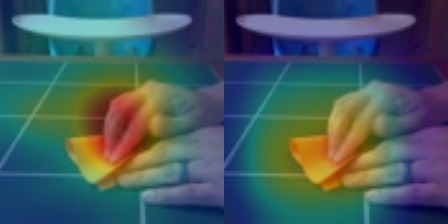} \\
			\includegraphics[scale=0.15]{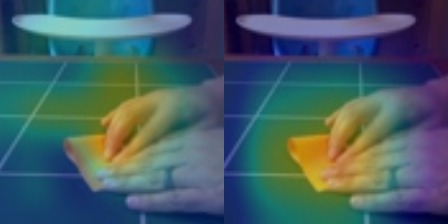} \\
			\includegraphics[scale=0.15]{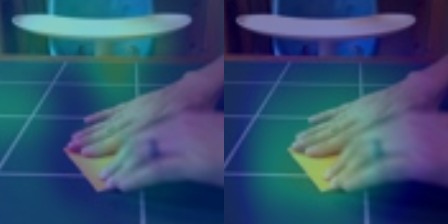} \\
			\includegraphics[scale=0.15]{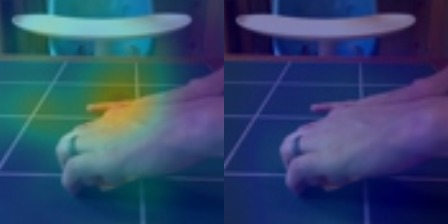} \\
			\includegraphics[scale=0.15]{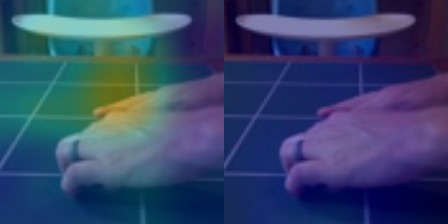} \\
			\includegraphics[scale=0.15]{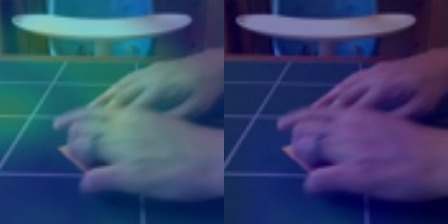} \\
			\includegraphics[scale=0.15]{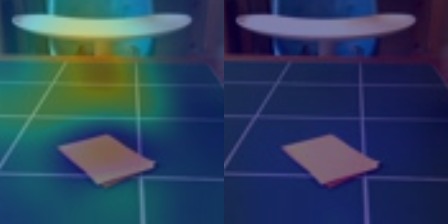} \\
			\includegraphics[scale=0.15]{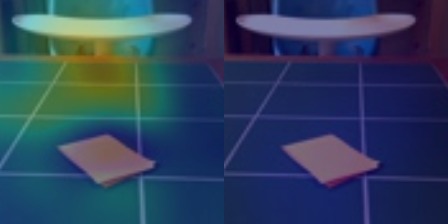}  
			\caption{}
		\end{subfigure} 
		\begin{subfigure}[b]{0.005\textwidth} 
			\raisebox{4in}{\rotatebox[origin=c]{90}{\texttt{plugging something into something but pulling it right out as you remove your hand}}}
		\end{subfigure} \hskip -7mm
		\begin{subfigure}[b]{0.24\textwidth}
			\centering
			\includegraphics[scale=0.15]{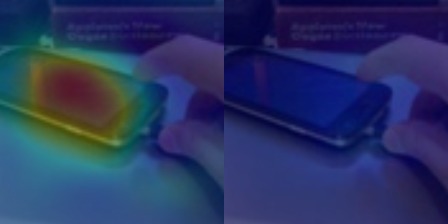} \\
			\includegraphics[scale=0.15]{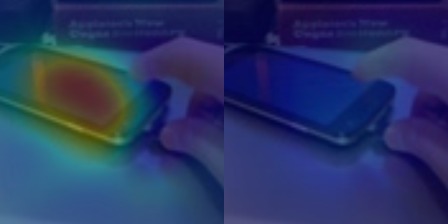} \\
			\includegraphics[scale=0.15]{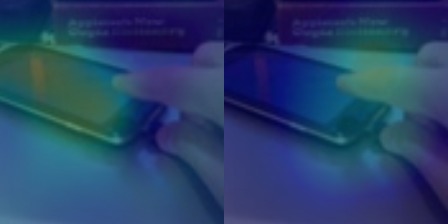} \\
			\includegraphics[scale=0.15]{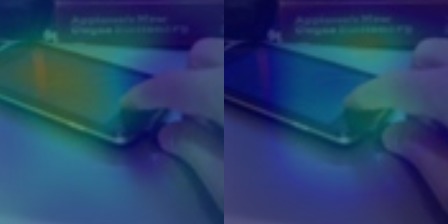} \\
			\includegraphics[scale=0.15]{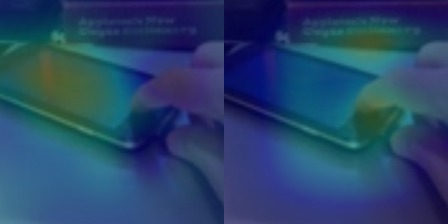} \\
			\includegraphics[scale=0.15]{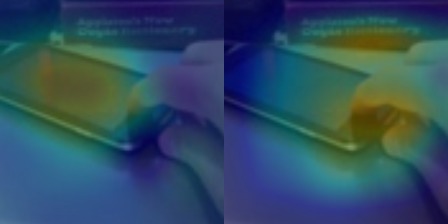} \\
			\includegraphics[scale=0.15]{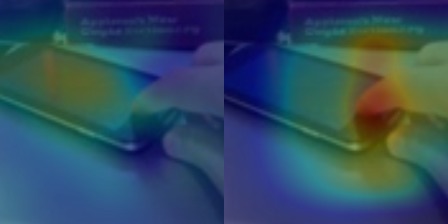} \\
			\includegraphics[scale=0.15]{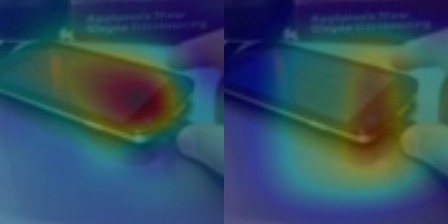} \\
			\includegraphics[scale=0.15]{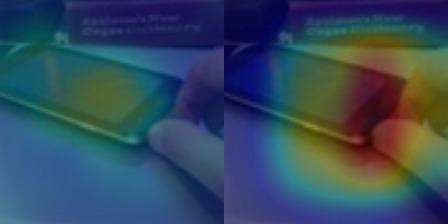} \\
			\includegraphics[scale=0.15]{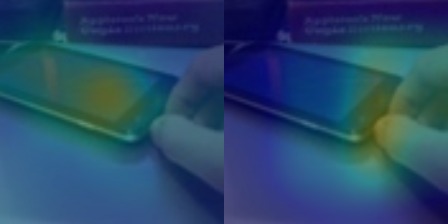} \\
			\includegraphics[scale=0.15]{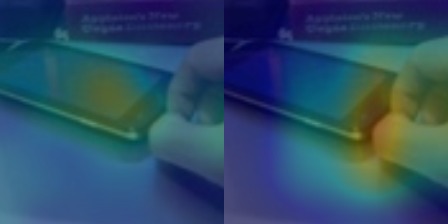} \\
			\includegraphics[scale=0.15]{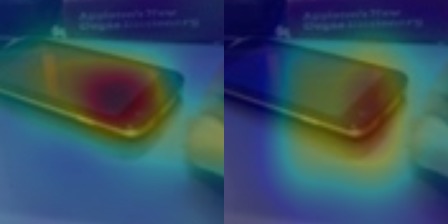} \\
			\includegraphics[scale=0.15]{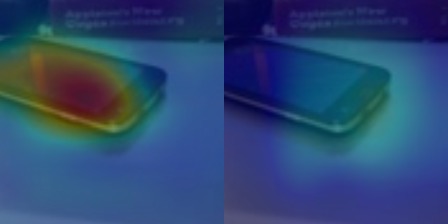} \\
			\includegraphics[scale=0.15]{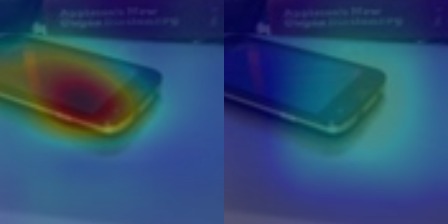} \\
			\includegraphics[scale=0.15]{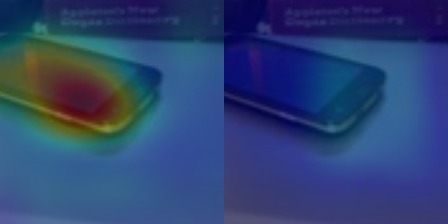} \\
			\includegraphics[scale=0.15]{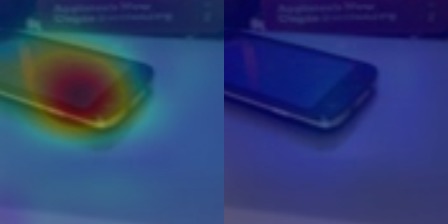}  
			\caption{}
		\end{subfigure}
		\begin{subfigure}[b]{0.005\textwidth} 
			\raisebox{4in}{\rotatebox[origin=c]{90}{\texttt{putting something similar to other things that are already on the table}}}
		\end{subfigure} \hskip -7mm
		\begin{subfigure}[b]{0.24\textwidth}
			\centering
			\includegraphics[scale=0.15]{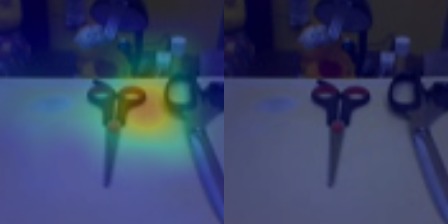} \\
			\includegraphics[scale=0.15]{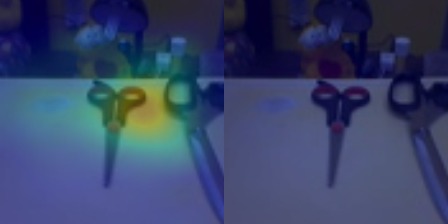} \\
			\includegraphics[scale=0.15]{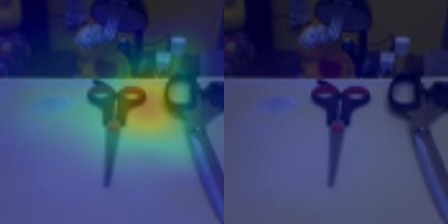} \\
			\includegraphics[scale=0.15]{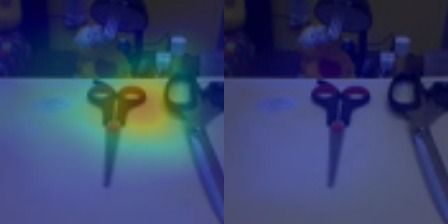} \\
			\includegraphics[scale=0.15]{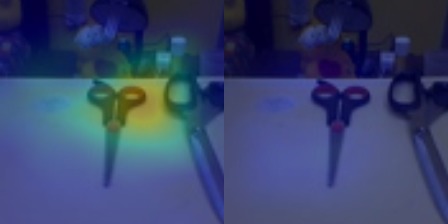} \\
			\includegraphics[scale=0.15]{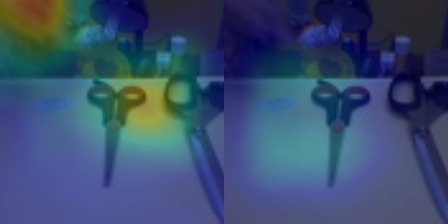} \\
			\includegraphics[scale=0.15]{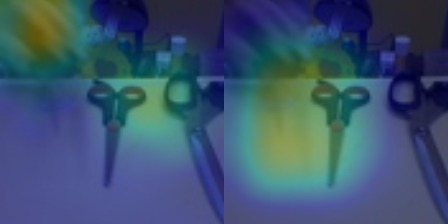} \\
			\includegraphics[scale=0.15]{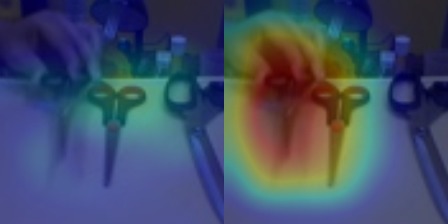} \\
			\includegraphics[scale=0.15]{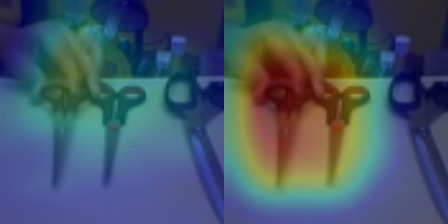} \\
			\includegraphics[scale=0.15]{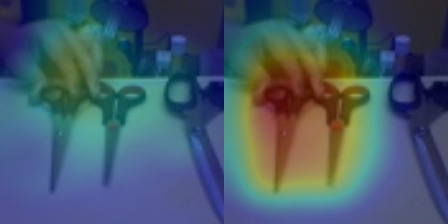} \\
			\includegraphics[scale=0.15]{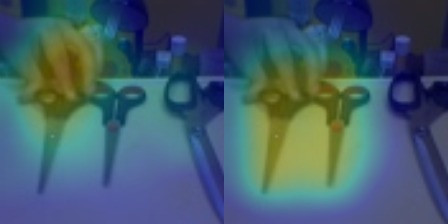} \\
			\includegraphics[scale=0.15]{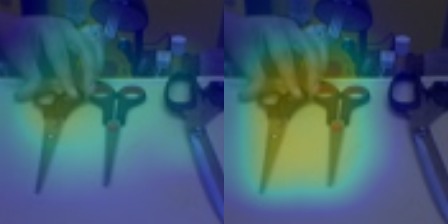} \\
			\includegraphics[scale=0.15]{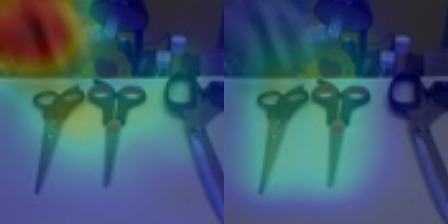} \\
			\includegraphics[scale=0.15]{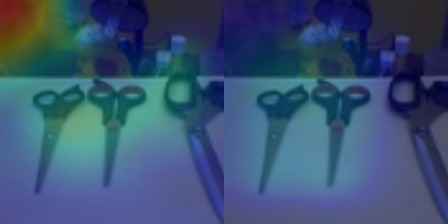} \\
			\includegraphics[scale=0.15]{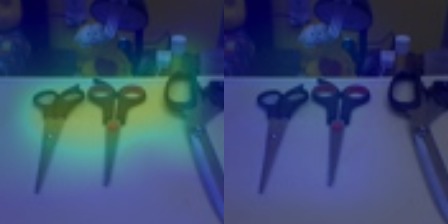} \\
			\includegraphics[scale=0.15]{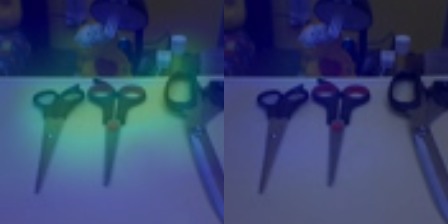}  
			\caption{}
		\end{subfigure}
		\caption{Saliency tubes generated by TSN (left) and GSM (right) on sample videos taken from the validation set of Something Something-V1 dataset. Action labels are shown as text on columns.
			}
		\label{fig:cam2}
	\end{figure*}

We show `visual explanations' for the decisions made by GSM. We use the approach of saliency tubes~\cite{saliency} for generating the visualizations. In this approach, the frames and their corresponding regions that are used by the model for making a decision are visualized in a form of saliency map. Figs.~\ref{fig:cam1} and~\ref{fig:cam2} compare the saliency tubes generated by the TSN baseline and the proposed GSM approach on sample videos from the validation set of Something Something-V1 dataset. We use the models with BNInception backbone trained using 16 frames for generating the visualizations. Each column in the figures show the 16 frames that are applied as input to the respective networks with the saliency tubes overlaid on top. We show TSN on the left side and GSM on the right side. The classes that improved the most by plugging in GSM on TSN are chosen for visualization. These classes require strong temporal reasoning for understanding the action. From the figures, we can see that TSN focuses on the objects present in the video irrespective of where and when the action takes place, while {\em GSM enables temporal reasoning by focusing on the active object(s) where and when an action is taking place}. For example, in Fig.~\ref{fig:put_object}, an example from the class \verb+putting something in front of something+, TSN focuses on the object that is present in the scene, the pen in the first few frames and the cup in the later frames. On the other hand, GSM makes the decision from the frames where the cup is introduced into the video. Similarly, in the example from the class \verb+taking one of many similar things on+ \verb+the table+ shown in Fig.~\ref{fig:remove}, TSN is focusing on the object, the matchbox, in all the frames while GSM makes the decision based on those frames where the action is taking place.

